\documentclass{article}
\usepackage{spconf,amsmath,graphicx}

\usepackage{booktabs}       
\usepackage{amsfonts}       
\usepackage{nicefrac}       
\usepackage{microtype}      
\usepackage{amsmath}
\usepackage{graphicx}
\usepackage{verbatim}
\usepackage{graphicx}
\usepackage{caption}
\usepackage{floatrow}
\usepackage{subfig}
\usepackage{breqn}
\usepackage{tabularx}

\usepackage[detect-all]{siunitx}
\usepackage{array}
\usepackage{adjustbox, multirow}

\newcolumntype{M}[1]{>{\centering\arraybackslash}m{#1}}

\title{Tensorizing Generative Adversarial Nets}
\name{Xingwei~Cao$^\dagger$, Xuyang~Zhao$^{\dagger \ddagger}$ and Qibin~Zhao$^\dagger$}
\address{$^\dagger$ RIKEN Center for Advanced Intelligence Project, Tokyo, Japan\\
	      $^\ddagger$ Saitama Institute of Technology, Saitama, Japan\\
	      \texttt{\{xingwei.cao,xuyang.zhao,qibin.zhao\}{@}riken.jp}}

\begin{document}
%
\maketitle
\begin{abstract}
Generative Adversarial Network (GAN) and its variants exhibit state-of-the-art performance in the class of generative models. To capture higher-dimensional distributions, the common learning procedure requires high computational complexity and a large number of parameters. The problem of employing such massive framework arises when deploying it on a platform with limited computational power such as mobile phones. In this paper, we present a new generative adversarial framework by representing each layer as a tensor structure connected by multilinear operations, aiming to reduce the number of model parameters by a large factor while preserving the generative performance and sample quality. To learn the model, we employ an efficient algorithm which alternatively optimizes both discriminator and generator. Experimental outcomes demonstrate that our model can achieve high compression rate for model parameters up to $35$ times when compared to the original GAN for MNIST dataset.
\end{abstract}
\section{Introduction}\label{intro}

Generative Adversarial Networks demonstrate state-of-the-art performance within the class of generative models \cite{goodfellow2014generative}. The success of GAN is accomplished not only by algorithmic advance but also by the recent growth of computational capacity. For example, state-of-the-art generative adversarial models such as \cite{karras2017progressive, zhang2017stackgan} utilize enormous computational resource by constructing complex models with a large number of model parameters and train them on powerful Graphics Processing Units (GPUs). 

A critical problem that accompanies with such dependency on powerful computational system arises when deploying the large-scale generative frameworks on platforms with limited computational power (\emph{e.g.} tablets, smartphones). For instance, mobile devices such as smartphones can only carry somewhat limited computational system due to its hardware design. Although unsupervised learning (especially generative adversarial learning) could improve the capability and functionality of mobile devices substantially, the trend of employing massive computational resource seems not profiting the usage of such generative framework for the general public.

The necessity of such complex model is partially attributed to the multi-dimensionality of datasets. Natural datasets often possess multi-modal structure, and among a plethora of such available datasets, images are often the subject of generative learning frameworks \cite{goodfellow2014generative, karras2017progressive, zhang2017stackgan}. As it becomes necessary to learn such high-dimensional dataset, models with a large number of parameters have been employed.

\begin{figure}[t]
\small
  \centering
  \includegraphics[width=8.5cm, height=3.5cm]{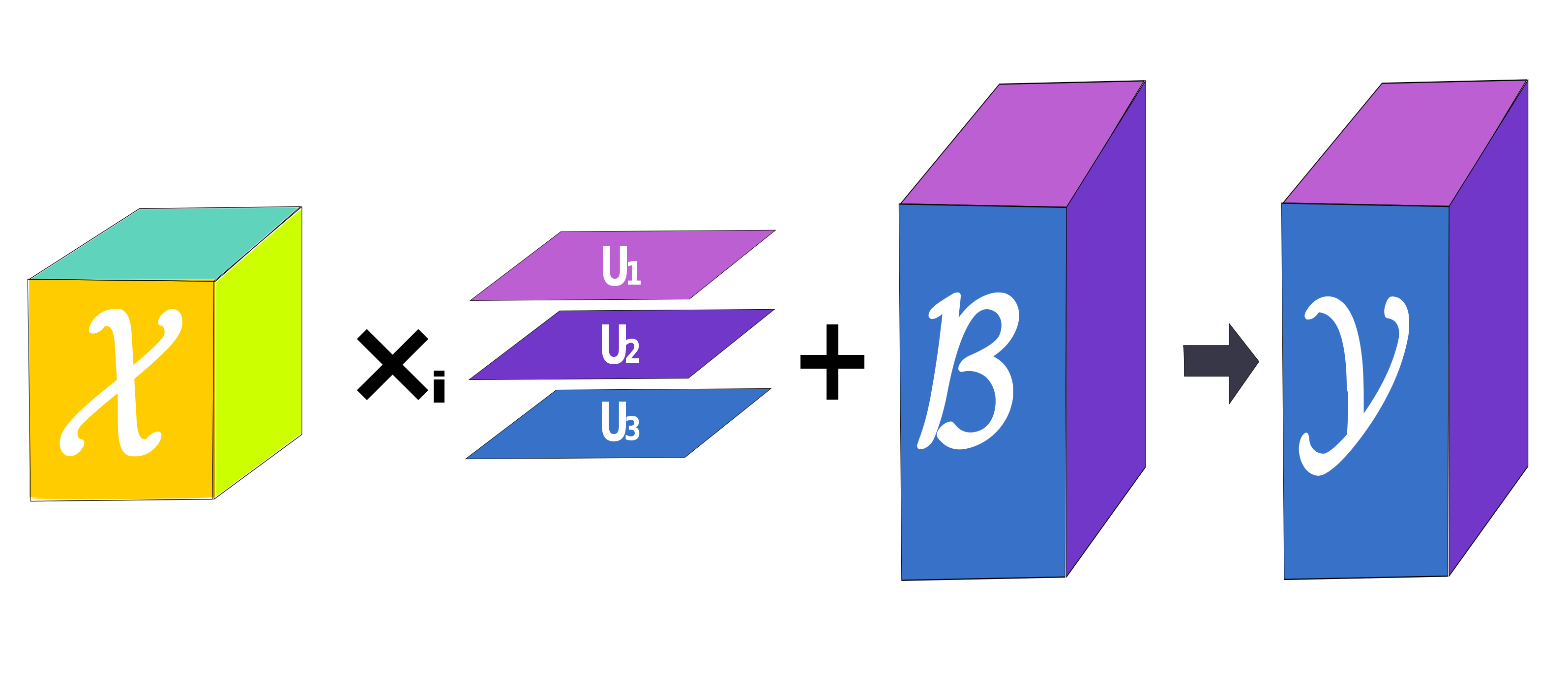}

  \caption{ \footnotesize Best viewed in color. Visualization of a three-way tensor layer. Given an input three-way tensor $\mathcal{X}\in\mathbb{R}^{I\times J\times K}$ and three weight matrices $\mathbf{U}_1\in\mathbb{R}^{L\times I}$, $\mathbf{U}_2\in\mathbb{R}^{M\times J}$ and $\mathbf{U}_3\in\mathbb{R}^{N\times K}$, the output is $\mathcal{Y}\in\mathbb{R}^{L\times M\times N}$. We call the resulting tensor $\mathcal{Y}$ after an activation a tensor layer and $\mathbf{U}_1$, $\mathbf{U}_2$ and $\mathbf{U}_3$ weight matrices.} \label{fig:tensorlayer}
\end{figure}

Goodfellow~et.~al. adopt multilayer perceptron (MLP) to learn and classify such higher-dimensional datasets \cite{goodfellow2014generative}. Although MLPs can represent rich classes of functions, the major drawback of them are, 1)~the dense connection between layers requires a large number of parameters, leading to a limited applicability of the framework to common computational environment, and 2)~the vectorization operation leads to the loss of rich inter-modal information of the datasets. 

In this paper, we propose a new generative framework with the purpose of reducing the number of model parameters while maintaining the quality of generated samples. Our framework is inspired by the recent works of applying tensor methods to machine learning models \cite{novikov2015tensorizing, kossaifi2017tensor}. For illustration, Novikov et al. proposed to use low-rank Tensor-Train (TT) approximations for weight parameters, where it shows that employing such tensor approximations will lead to the reduction of the space complexity of a model \cite{novikov2015tensorizing}. Although applying TT decomposition to dense matrices demonstrates a large factor of compression rate \cite{novikov2015tensorizing}, finding optimal TT-ranks still remains to be a difficult problem. 

In our proposed framework, we compress the traditional affine transformation using tensor algebra in order to reduce the number of parameters of fully-connected layers. In particular, all hidden, input and output layers are represented as a tensor structure, not as a vector. By treating a multi-dimensional input without vectorization, our model aims to preserve its original multi-modal information while saving the space complexity of a model by a large factor. The usage of tensor algebra results in model compression while preserving multi-modal information of dataset.

We empirically demonstrate high compression rate of our proposed model with both benchmark and synthetic datasets. We compare our framework with models without tensorization, displaying that generative learning process could be accomplished with a smaller number of model parameters. In our experiment with a dataset of handwritten digits, we observed that our model achieves the compression rate of $40$ times while producing images with a comparable quality.

The rest of this paper is organized as follows. We start with the concise review of basic tensor arithmetics and generative adversarial nets in Section~\ref{sc:background}. Section~\ref{sc:tensorlayer} introduces the tensor layer. In Section~\ref{sc:training}, we present the swift applicability of state-of-the-art learning algorithms to our framework. Experimental results and details are shown in Section~\ref{sc:exp} followed by the conclusion in Section~\ref{sc:conclusion}.

\section{Background} \label{sc:background}

\subsection{Tensor algebra}  \label{subsc:tensor}
There are multiple tensor operations necessary to construct a tensor layer. In particular, the $n$-mode product, Kronecker product and Hadamard product are briefly reviewed in this section. For more comprehensive review of tensor arithmetics and notations, see \cite{kolda2009tensor}. 

The \emph{order} of a tensor is the number of dimensions. Vector, matrix and tensor with order three or higher are denoted by $\bf{a}$, $\bf{A}$ and $\mathcal{A}$ respectively. Given an $N$th-order tensor $\mathcal{X}\in\mathbb{R}^{I_{1}\times I_{2}\times\cdots\times I_{N}}$, its $(i_1, i_2, \ldots, i_N)$th entry is denoted by $\mathcal{X}_{i_{1}i_{2}\ldots i_{N}}$, where $i_{n}=1, 2,\ldots, I_{n}, \forall n\in[1,N]$. The notation $[I, J]$ denotes a set of integers ranged from $I$ to $J$ inclusive.

The mode-\emph{n} fibers are the vectors obtained by fixing every index but the \emph{n}-th index of a tensor. The mode-\emph{n} matricization or mode-\emph{n} unfolding of a tensor $\mathcal{X}$ is denoted by $\bf{X}_{(n)}$. It arranges the mode-n fibers to be the columns of the resulting matrix. Given matrices $\mathbf{A}$ and $\mathbf{B}$, both of same size $\mathbb{R}^{I\times J}$, their \emph{Hadamard product} (or component-wise product) is denoted by $A*B$. The resulting matrix is also of the size $I\times J$.

The $n$-mode product of a tensor $\mathcal{X}\in\mathbb{R}^{I_{1}\times I_{2}\times\cdots\times I_{N}}$ with a matrix $\mathbf{U}\in\mathbb{R}^{J\times I_{n}}$ is denoted by $\mathcal{X}\times_n \mathbf{U} \in \mathbb{R}^{I_{1}\times \cdots \times  I_{n-1} \times J \times I_{n+1} \times \cdots \times I_{N}}$ and an entry of the product is defined by

\begin{equation}
  (\mathcal{X}\times_n\mathbf{U})_{i_1\cdots i_{n-1}\, j \, i_{n+1}\cdots i_N} = \sum_{i_n =1}^{I_n} x_{i_1 i_2 \cdots i_N}\, u_{j i_n}.
\end{equation}

The \emph{Kronecker product} of matrices $\mathbf{A}\in\mathbb{R}^{I\times J}$ and $\mathbf{B}\in\mathbb{R}^{K\times L}$ is a matrix of size $IK\times JL$, denoted by $\mathbf{A}\otimes \mathbf{B}$. The product is defined by

\begin{equation}
   \mathbf{A}\otimes \mathbf{B} =  \begin{bmatrix}
  
    a_{11}B & a_{12}B & a_{13}B & \dots  & a_{1J}B \\
    a_{21}B & a_{22}B & a_{23}B & \dots  & a_{2J}B \\
    \vdots & \vdots & \vdots & \ddots & \vdots \\
    a_{I1}B & a_{I2}B & a_{I3}B & \dots  & a_{IJ}B
\end{bmatrix}.
\end{equation}

We use one of the properties of the Kronecker product in this paper. Given a tensor $\mathcal{X}\in\mathbb{R}^{I_{1}\times I_{2}\times\cdots\times I_{N}}$ and a set of matrices denoted by $\mathbf{A}^{(n)}\in \mathbb{R}^{J_n\times I_n}$ for $n\in[1, N]$,

\begin{equation}\label{eq:property}
\begin{aligned}
\mathcal{Y} &= \mathcal{X}\times_1\mathbf{A}^{(1)}\times_2\mathbf{A}^{(2)}\times_3 \cdots \times_N\mathbf{A}^{(N)} \\
& \iff \mathbf{Y}_{(n)} = \mathbf{A}^{(n)}\mathbf{X}_{(n)} \left( \prod_{\substack{i=1 \\ i\neq n}}^{N} \otimes \mathbf{A}^{(i)} \right)^T.
\end{aligned}
\end{equation}

\subsection{Generative adversarial nets}  \label{subsc:gan}
Generative adversarial nets consist of two components called generator and discriminator, which usually are represented by MLPs. The task of the discriminator is to correctly identify whether the input belongs to the real data distribution $p_{data}$ or the model distribution $p_{model}$. Given a prior $z\sim p_z$, the generator tries to produce indistinguishable samples to deceive the discriminator. By alternatively training discriminator and generator, GAN aims to implicitly learn two distributions $p_{data}$ and $p_{model}$. Due to the space limitation, we abbreviate the details of GAN trainings. For more detailed explanation of GANs, we refer readers to \cite{goodfellow2014generative}.

\section{Tensor layer}  \label{sc:tensorlayer}

In this section, we introduce the \emph{Tensor layer} using tensor algebra we reviewed in Section~\ref{subsc:tensor}. A \emph{Tensor layer} replaces the traditional affine transformation between layers of MLP, which forms GANs, with multilinear affine transformations. When we train an MLP model, it is common to flatten the inputs then feed those vectors to the network. Given such input vector $\bf{x}$, the traditional transformation for MLPs is defined as follows: 

\begin{equation}
\bf{y} = \text{$\sigma$} \left( \mathbf{W} \bf{x} + \bf{b} \right).
\end{equation}

Tensor layer treats the input without vectorization. Given an N-way tensor $\mathcal{X}\in\mathbb{R}^{I_1 \cdots \times I_N}$ as an input, rather having one matrix $\mathbf{W}$, we establish the transformation between two tensor layers by applying mode product operations between the input tensor and \emph{weight matrices} $\mathbf{U}_i$ for $i\in[1,N]$. We thereafter add a \emph{bias tensor} $\mathcal{B}\in\mathbb{R}^{J_1 \times \cdots \times J_N}$ to the product to form a tensor layer. The transformation from $\mathcal{X}$ to the tensor layer $\mathcal{Y}\in\mathbb{R}^{J_1 \times \cdots \times J_N}$ is formulated as:

\begin{equation}
\mathcal{Y} = \text{$\sigma$} \left( \mathcal{X}\times_1{\mathbf{U}_1}\times_2{\mathbf{U}_2}\times_3 \cdots \times_N \mathbf{U}_N+\mathcal{B} \right)
\end{equation}
where $\mathbf{U}_i\in\mathbb{R}^{J_i \times I_i}$ for $i\in [1,N]$. A visualization for a third order input is provided in Figure~\ref{fig:tensorlayer}.

Tensor layers could be interpreted as Tucker Decomposition, which factorizes a higher-order tensor into a core tensor and factor matrices  \cite{tucker1966some, kolda2009tensor}. If we treat the input tensor $\mathcal{X}$ as such core tensor, the weight matrices $\mathbf{U}_i$ for $i\in [1,N]$ could be interpreted as factor matrices.

\section{Training FT-Nets} \label{sc:training}

In this paper, networks consisting of tensor layers are referred as \emph{FT-Nets}. A generative adversarial nets consisting of FT-NETs are called \emph{TGAN}. In this section, we demonstrate that the gradient-based back-propagation algorithms are swiftly applicable to \emph{FT-Nets}.

Given a FT-Net with an input tensor $\mathcal{X}\in\mathbb{R}^{I_1\times \cdots \times I_N \times C}$ and two tensor layers $\mathcal{O}_1\in\mathbb{R}^{J_1 \ \times \cdots \times J_N \times C}$ and $\mathcal{O}_2\in\mathbb{R}^{K_1 \times \cdots \times K_N \times C}$, we can formulate such FT-Net as follows;

\begin{dmath}
\mathcal{O}_1 = g\left( \mathcal{H}_1 \right) = g\left( \mathcal{X} \times_1 \mathbf{W}_1 \times_2  \mathbf{W}_2 \times_3  \cdots \times_N \mathbf{W}_N + \mathcal{B}_1\right)
\end{dmath}

\begin{dmath}
\mathcal{O}_2 = g \left( \mathcal{H}_2 \right) = g \left( \mathcal{O}_1 \times_1 \mathbf{U}_1 \times_2  \mathbf{U}_2 \times_3  \cdots \times_N \mathbf{U}_N + \mathcal{B}_2 \right)
\end{dmath}

where $\mathbf{W}_i\in \mathbb{R}^{J_i \times I_i}$ for $i \in [1,N]$ and $\mathbf{U}_i\in \mathbb{R}^{K_i \times J_i}$ for $i \in [1,N]$. The function $g(\mathcal{A})$ is a component-wise activation function applied to a tensor $\mathcal{A}$.  Note that the last mode of the input tensor $\mathcal{X}$ denotes the size of each batch $C$, thus we do not apply any mode product to the $N+1$ th mode when performing the multilinear operation to $\mathcal{X}$. 

Given an output layer $\mathcal{H}_2$, the gradient of $\mathcal{H}_2$ with respect to each weight matrix $\mathbf{U}_i$ is derived using Eq.~\eqref{eq:property} $\forall i \in [1,N]$;

\begin{dmath}
\frac{\partial \mathcal{H}_2}{\partial \mathbf{U}_i}  = \frac{\partial}{\partial \mathbf{U}_i} \left( \mathcal{O}_1 \times_1 \mathbf{U}_1 \times_2  \mathbf{U}_2 \times_3  \cdots \times_N \mathbf{U}_N + \mathcal{B}_1 \right) = \frac{\partial}{\partial \mathbf{U}_i} \left( \mathbf{U}_i \mathcal{O}_{1(i)} \left( \prod_{\substack{j=1 \\ j\neq i}}^{N} \otimes \mathbf{U}_j \ \right)^{T} +\mathcal{B}_{1(i)} \right)
\end{dmath}

The gradient of $\mathcal{H}_2$ with respect to $\mathcal{O}_1$ can be derived as 
\begin{dmath}
\frac{\partial \mathcal{H}_2}{\partial \mathcal{O}_1} = \frac{\partial}{\partial \mathcal{O}_1} \left(\mathcal{O}_1 \times_1 \mathbf{U}_1 \times_2  \mathbf{U}_2 \times_3  \cdots \times_N \mathbf{U}_N + \mathcal{B}_2 \right) = \mathbf{U}_N \otimes \mathbf{U}_{N-1} \otimes \cdots \otimes \mathbf{U}_1
\end{dmath}

We can derive the gradient of a loss function $f$ with respect to $\mathbf{W}_i$ for $i \in [1,N]$ as follows.

\begin{dmath}
\frac{\partial f}{\partial \mathbf{W}_i} = \Biggl(\biggl( \left( \frac{\partial f}{\partial \mathcal{O}_2} * \frac{\partial \mathcal{O}_2}{\partial \mathcal{H}_2} \right) \frac{\partial \mathcal{H}_2}{\partial \mathcal{O}_1}  \biggr) * \frac{\partial \mathcal{O}_1}{\partial \mathcal{H}_1}  \Biggr) \frac{\partial \mathcal{H}_1}{\partial \mathbf{W}_i} = \left(\left( \left( \frac{\partial f}{\partial \mathcal{O}_2} * \frac{\partial \mathcal{O}_2}{\partial \mathcal{H}_2} \right) \prod_{j=N}^{1} \otimes \mathbf{U}_j \right) * \frac{\partial \mathcal{O}_1}{\partial \mathcal{H}_1}  \right) \mathcal{O}_{1(i)} \left( \prod_{\substack{j=1 \\ j\neq i}}^{N} \otimes \mathbf{W}^{(j)} \right)^{T}
\end{dmath}

Given a neural network with three fully-connected layers of size $I = \prod_{i=1}^{N}I_i$, $J =\prod_{i=1}^{N} J_i$ and $K =\prod_{i=1}^{N} K_i$, the number of model parameters in the network is $J\left(I+K\right) + J + K$. We can simply use such factorizations of $I$, $J$ and $K$ to formulate a FT-Net. The number of model parameters of the FT-Net is $\sum_{i = 1}^{N} \left(I_i J_i + J_i K_i \right)+J + K$. It is easily observable that the difference in the number of parameters between tensorized and un-tensorized networks can grow exponentially as the order of the input tensor increases.

\section{Experiments} \label{sc:exp}

In this section we empirically demonstrate the compressive power of our framework using both synthetic and real dataset. All Hyper-parameters and architectural details for every experiment are available at https://github.com/xwcao/TGAN.

\subsection{MNIST} \label{mnist}

\begin{figure}[t]
\centering
\begin{tabular}{c   M{1.31cm} M{1.31cm} M{1.31cm} M{1.31cm}  M{1.31cm} c }
 \hline
 
 {\footnotesize \multirow{ 2}{*}{Model}} & \multicolumn{5}{c}{ {\footnotesize Iterations}} &  {\footnotesize \multirow{ 2}{*}{\# Params}}\\

 & {\footnotesize 10k} & {\footnotesize 20k}  & {\footnotesize 30k}  & {\footnotesize 40k}  & {\footnotesize 50k} & \\

  \hline
  
  {\footnotesize TGAN}
  &\includegraphics[width=1.31cm, height=1.31cm]{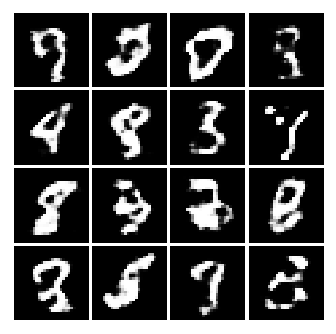}
  &\includegraphics[width=1.31cm, height=1.31cm]{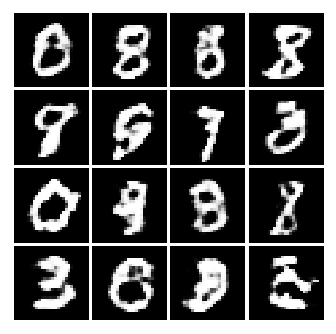}
  &\includegraphics[width=1.31cm, height=1.31cm]{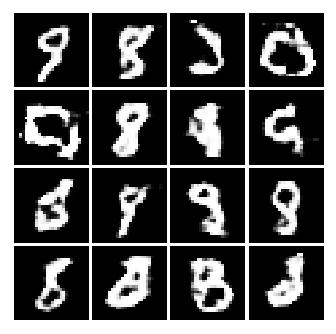}
  &\includegraphics[width=1.31cm, height=1.31cm]{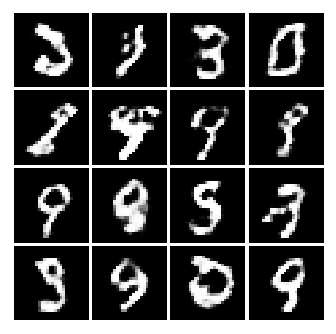}
  &\includegraphics[width=1.31cm, height=1.31cm]{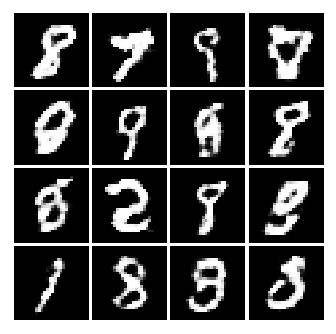}
  &{\footnotesize $12$k}\\
    
  
  {\footnotesize GAN$^{(1)}$}
  &\includegraphics[width=1.31cm, height=1.31cm]{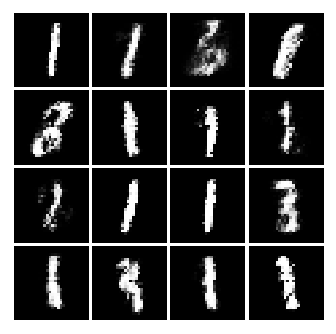}
  &\includegraphics[width=1.31cm, height=1.31cm]{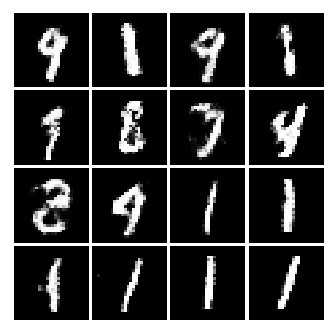}
  &\includegraphics[width=1.31cm, height=1.31cm]{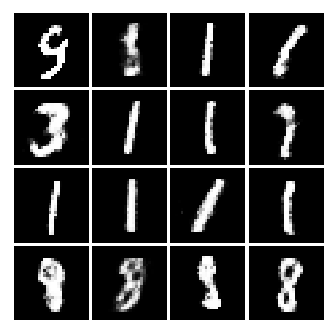}
  &\includegraphics[width=1.31cm, height=1.31cm]{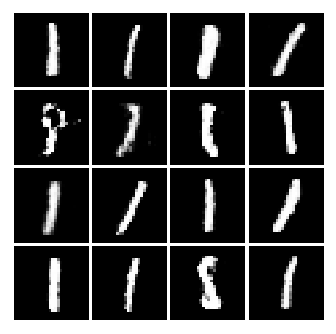}
  &\includegraphics[width=1.31cm, height=1.31cm]{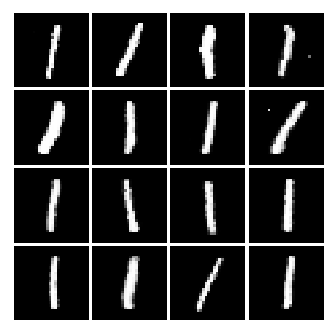}
  &{\footnotesize $429$k}\\
  
  
  {\footnotesize GAN$^{(2)}$}
  &\includegraphics[width=1.31cm, height=1.31cm]{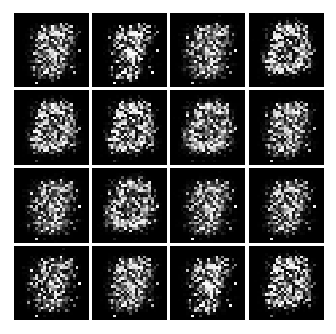}
  &\includegraphics[width=1.31cm, height=1.31cm]{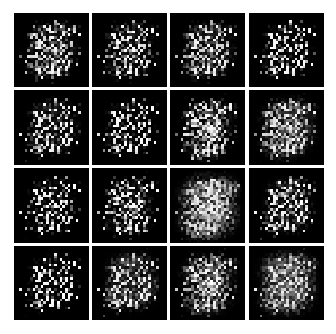}
  &\includegraphics[width=1.31cm, height=1.31cm]{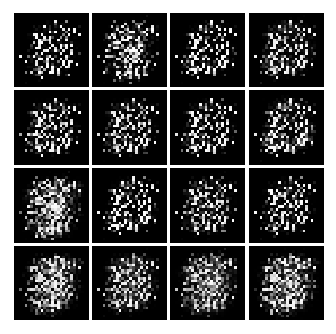}
  &\includegraphics[width=1.31cm, height=1.31cm]{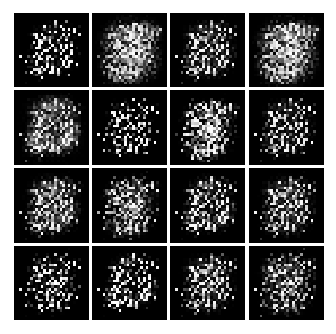}
  &\includegraphics[width=1.31cm, height=1.31cm]{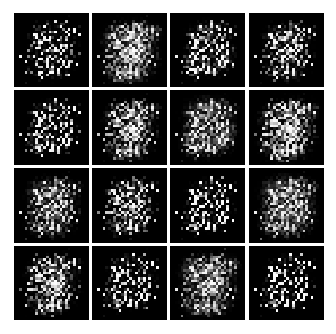}
  &{\footnotesize $12$k}\\

 \hline
  
  \end{tabular}
  \caption{ \footnotesize Comparison of MNIST samples generated by TGAN and GANs. We collected the samples in the figure without cherry-picking. We used the prior $z\sim \mathcal{U}(-1,1)$ to sample randomly for both TGAN and GANs. Our framework demonstrates its compressive power when compared to GAN$^{(1)}$ and GAN$^{(2)}$.}\label{fig:mnist}
\end{figure}

In this section, we present empirical comparisons of TGAN and GAN using images of handwritten digits \cite{lecun1998mnist}. We conducted experiments with TGAN and two GANs. The first GAN, namely GAN$^{(1)}$ has larger amount of parameters than TGAN, and GAN$^{(2)}$ has approximately same number of parameters as TGAN. See Figure~\ref{fig:mnist} for experimental outcomes. Although numerical evaluation of generative model is an open problem, we believe that quality of samples from TGAN and GAN$^{(1)}$ are comparable. We report the factor of the compression rate between TGAN and GAN$^{(1)}$ to be 35 times.

\captionsetup[subfloat]{labelformat=empty}
\def\arraystretch{1}\tabcolsep=0.1pt

\subsection{Synthetic Data} \label{subsc:synthetic}

\begin{figure}[t]
\centering
\begin{tabular}{c   M{0.95cm} M{0.95cm} M{0.95cm} M{0.95cm} M{0.95cm} M{0.95cm} M{0.95cm} }
 \hline
 
 {\footnotesize \multirow{ 2}{*}{Model}} & \multicolumn{6}{c}{ {\footnotesize Iterations}}  &  {\footnotesize \multirow{ 2}{*}{True}} \\

 & {\footnotesize 1k} & {\footnotesize 2k}  & {\footnotesize 4k}  & {\footnotesize 6k}  & {\footnotesize 8k} & {\footnotesize 10k} & \\

  \hline
  
  {\footnotesize TGAN}
  &\includegraphics[width=0.95cm, height=0.95cm]{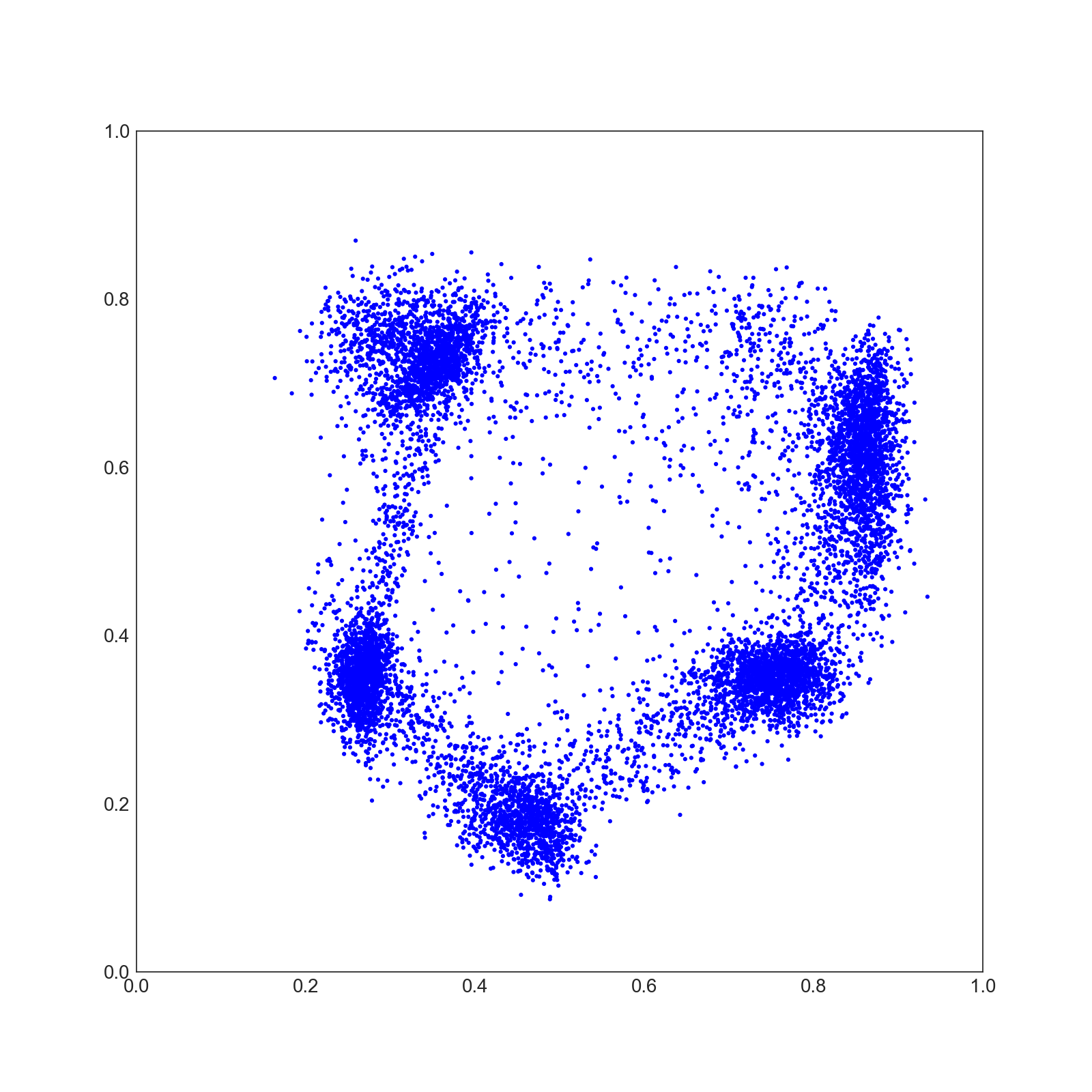}
  &\includegraphics[width=0.95cm, height=0.95cm]{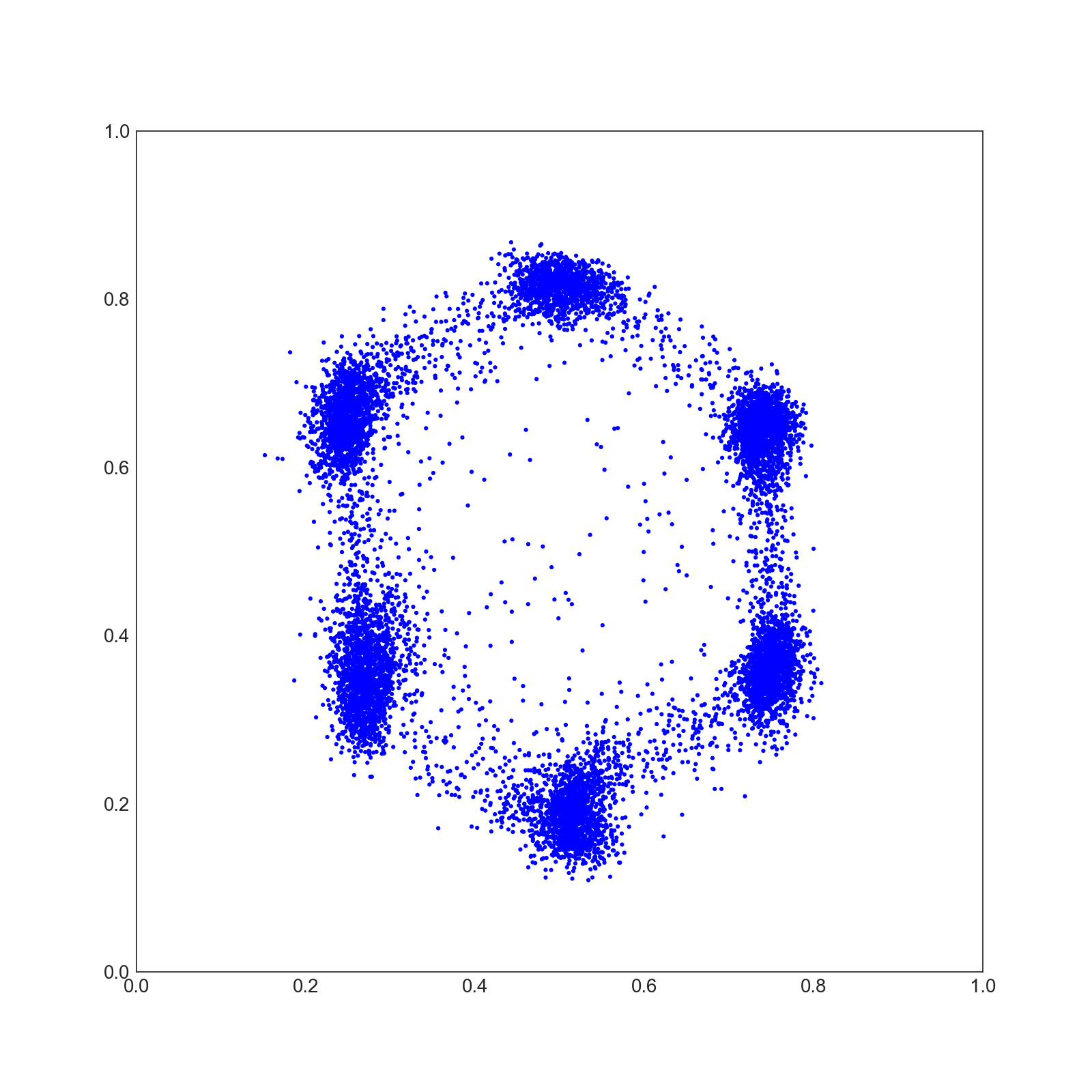}
  &\includegraphics[width=0.95cm, height=0.95cm]{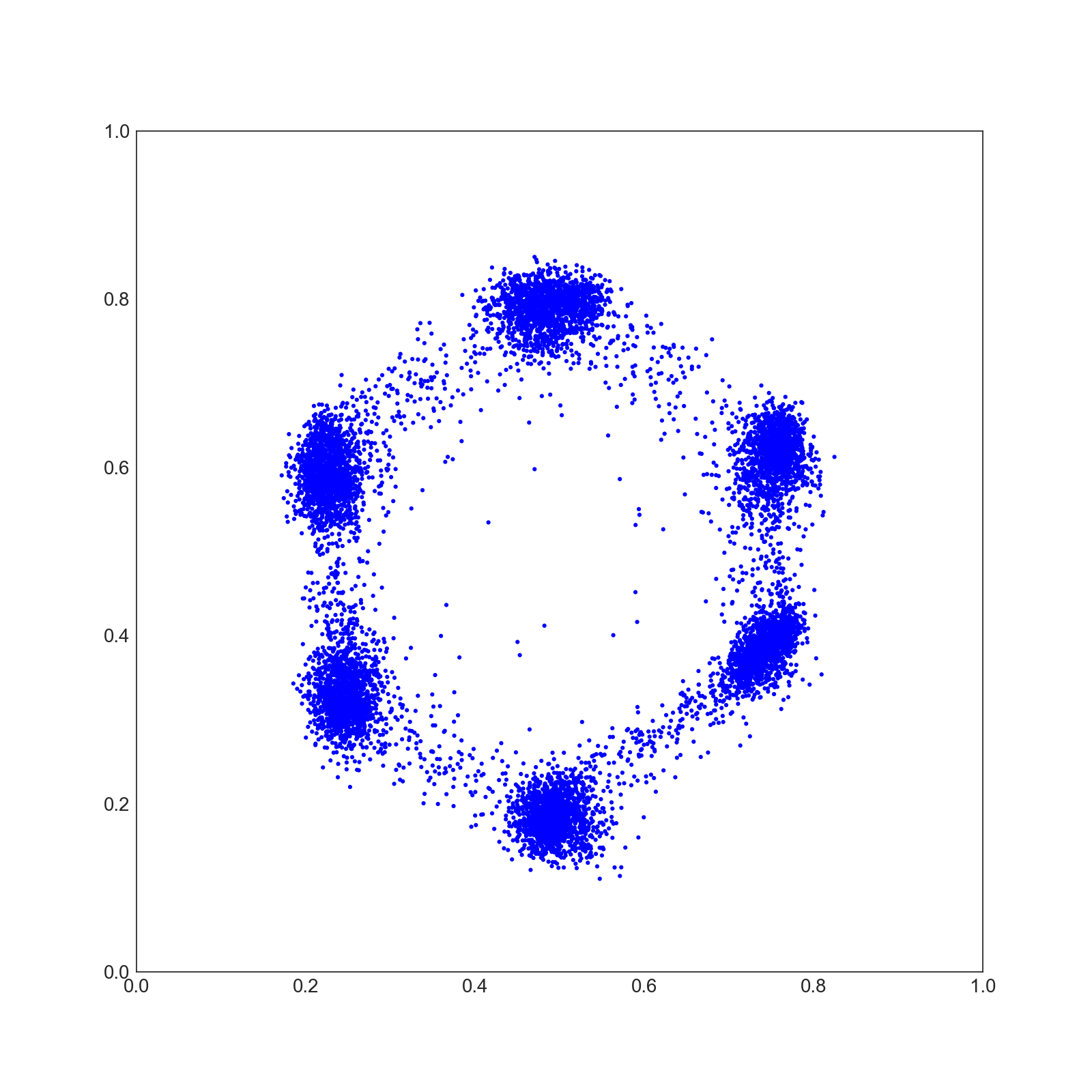}
  &\includegraphics[width=0.95cm, height=0.95cm]{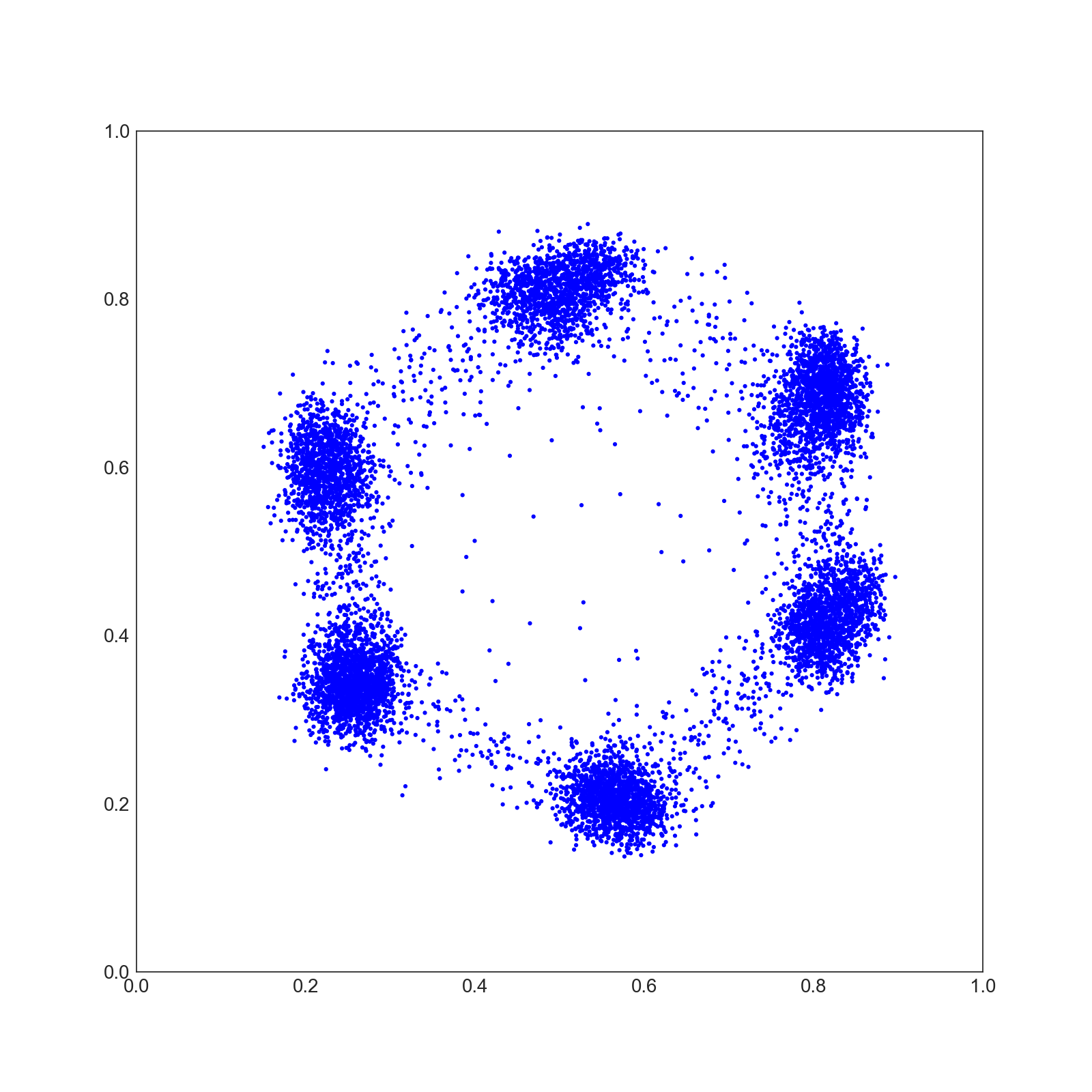}
  &\includegraphics[width=0.95cm, height=0.95cm]{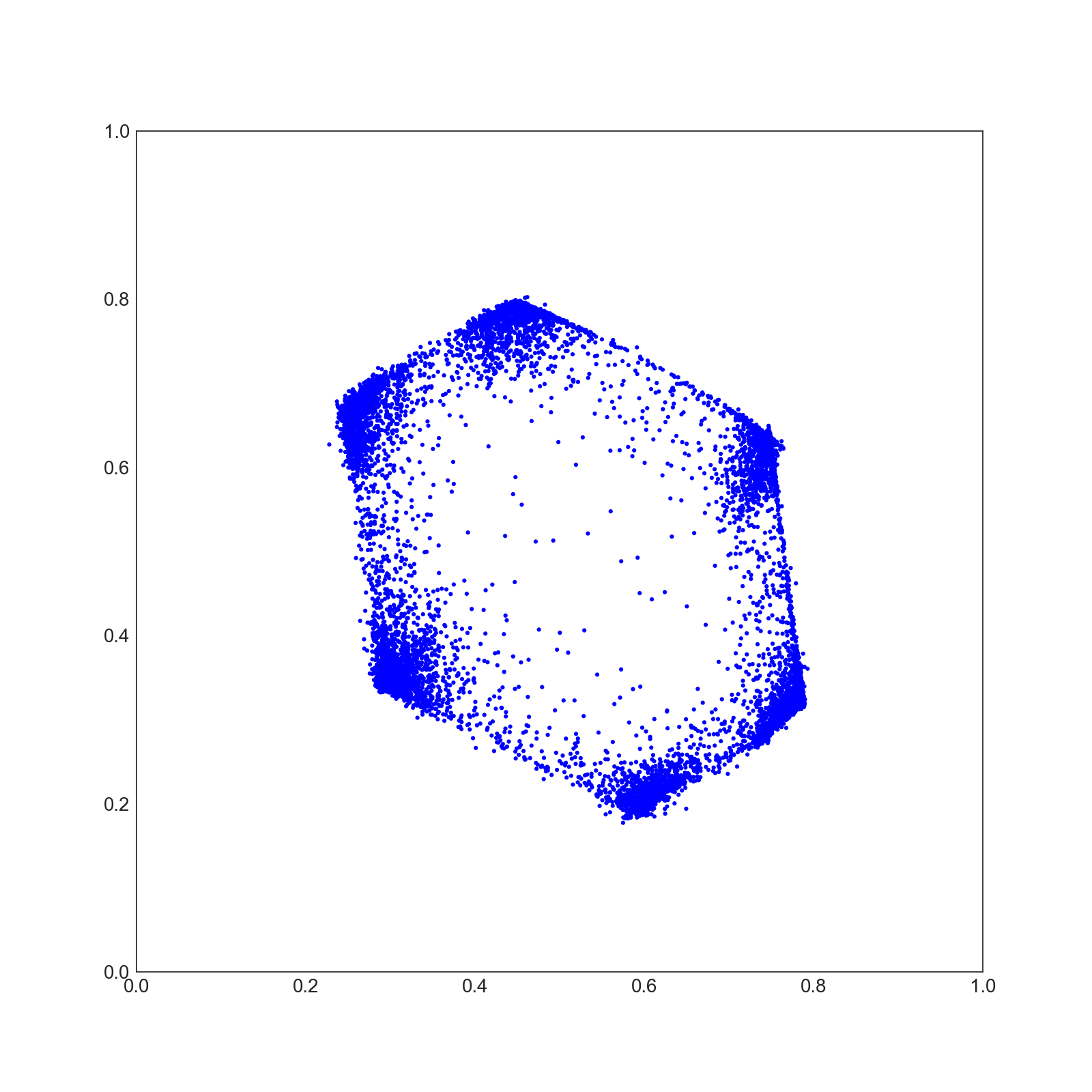}
  &\includegraphics[width=0.95cm, height=0.95cm]{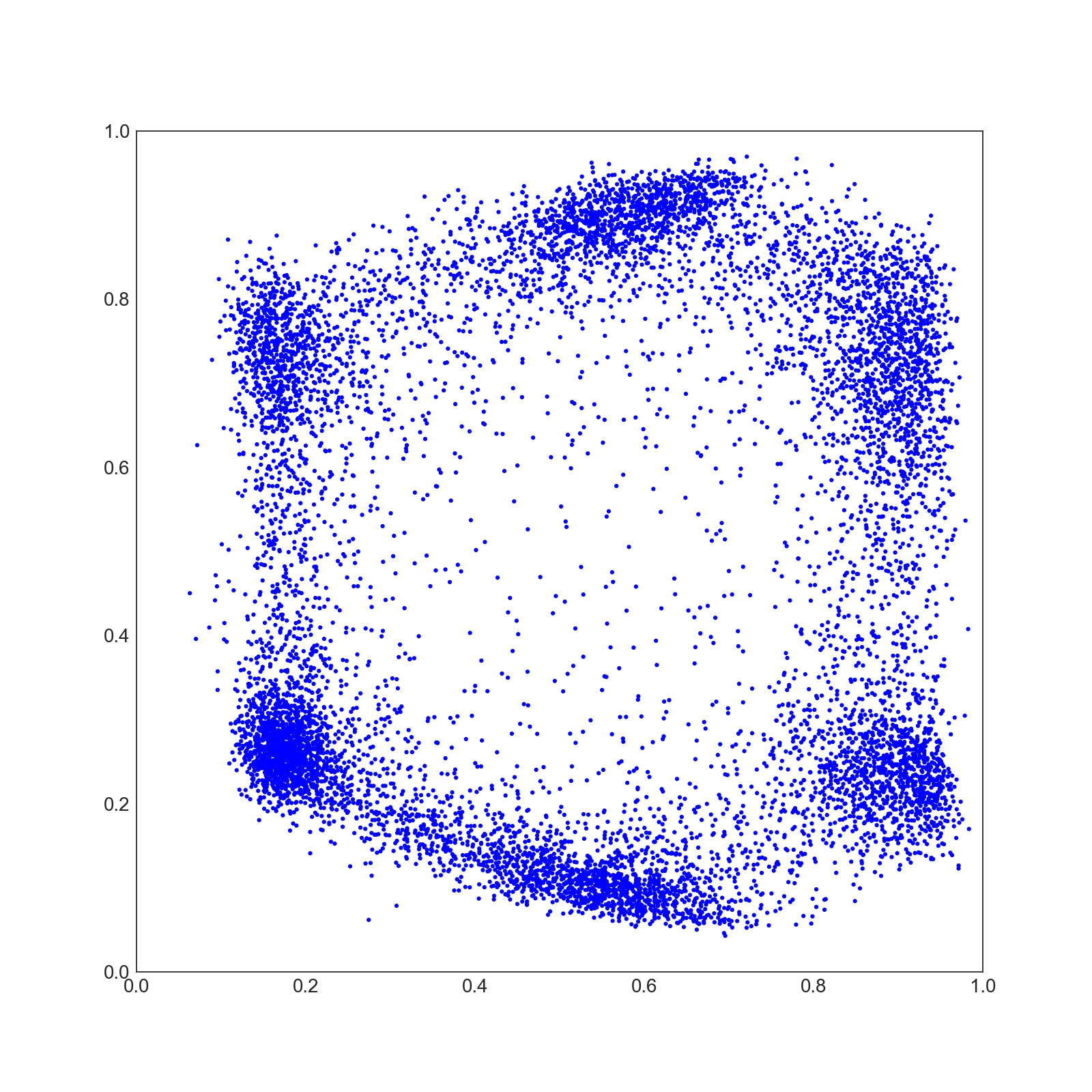}
  &\includegraphics[width=0.95cm, height=0.95cm]{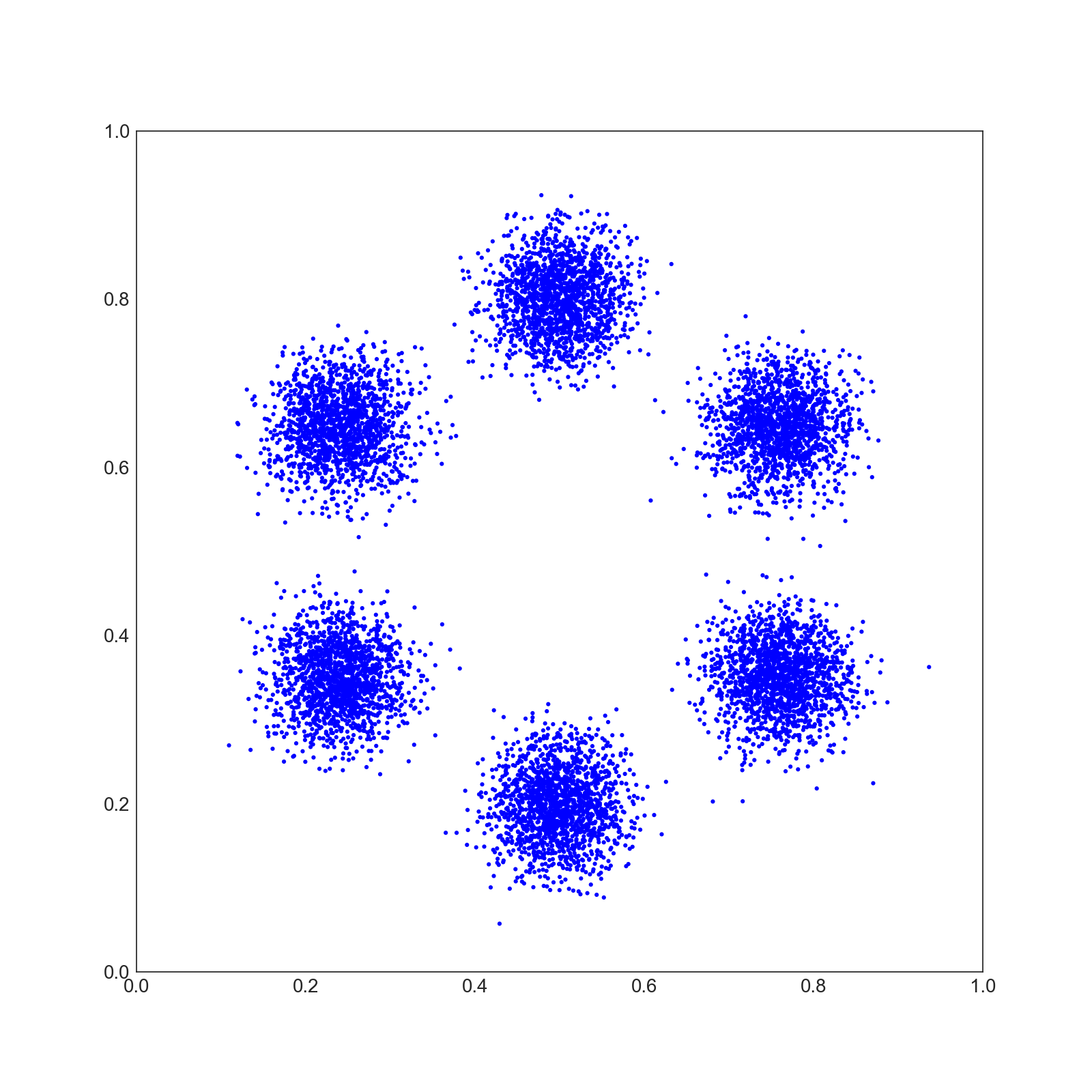}\\
    
  
  {\footnotesize GAN$^{(1)}$}
  &\includegraphics[width=0.95cm, height=0.95cm]{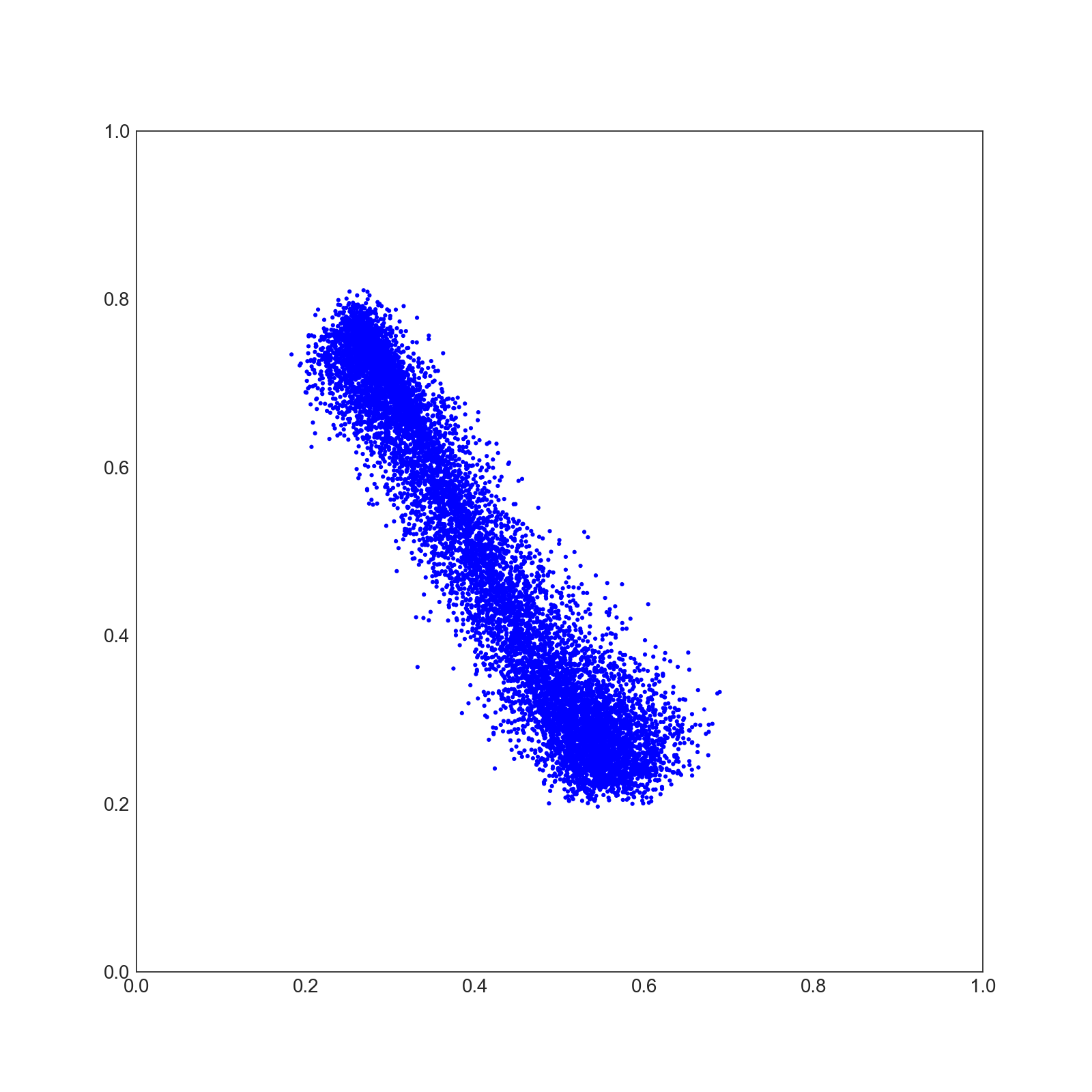}
  &\includegraphics[width=0.95cm, height=0.95cm]{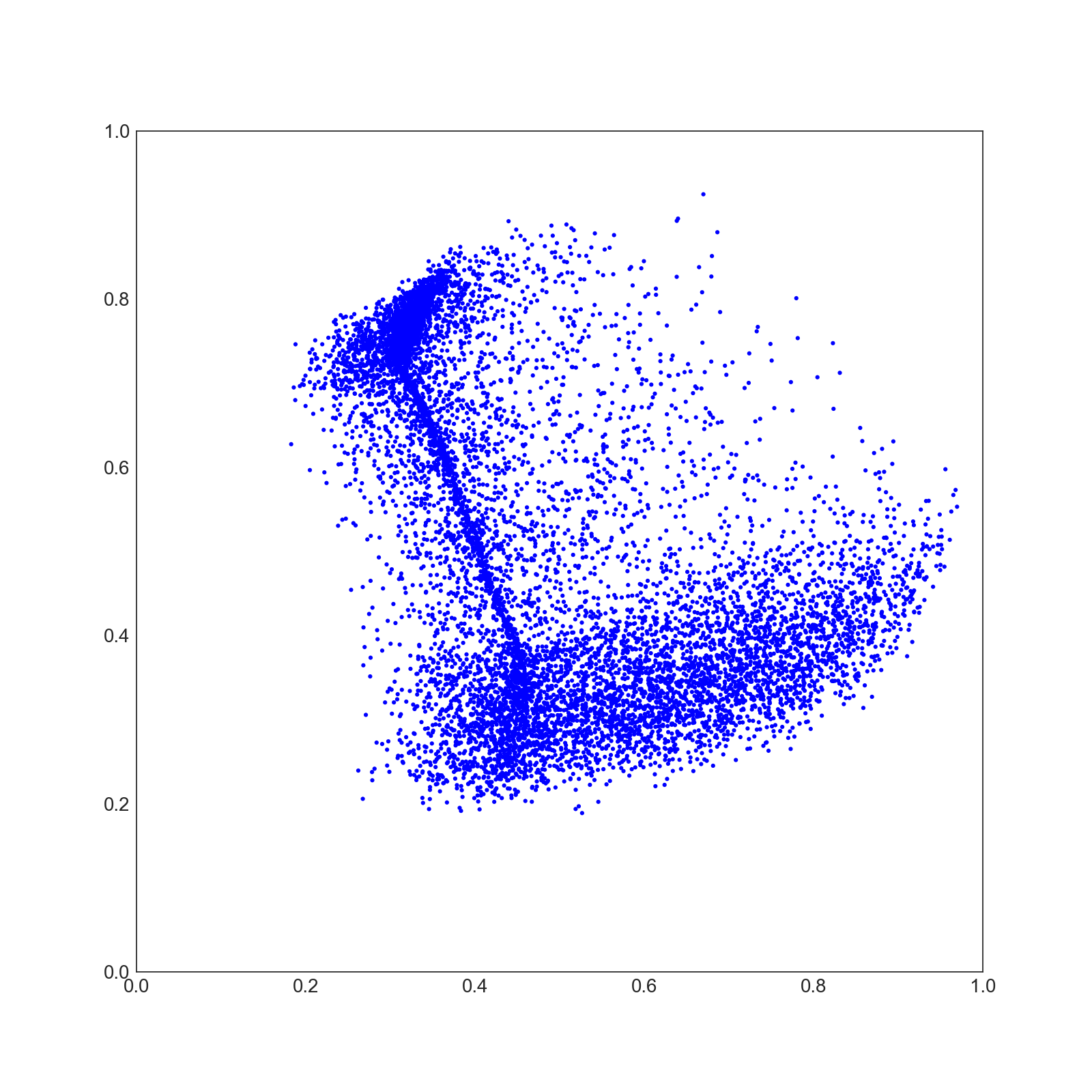}
  &\includegraphics[width=0.95cm, height=0.95cm]{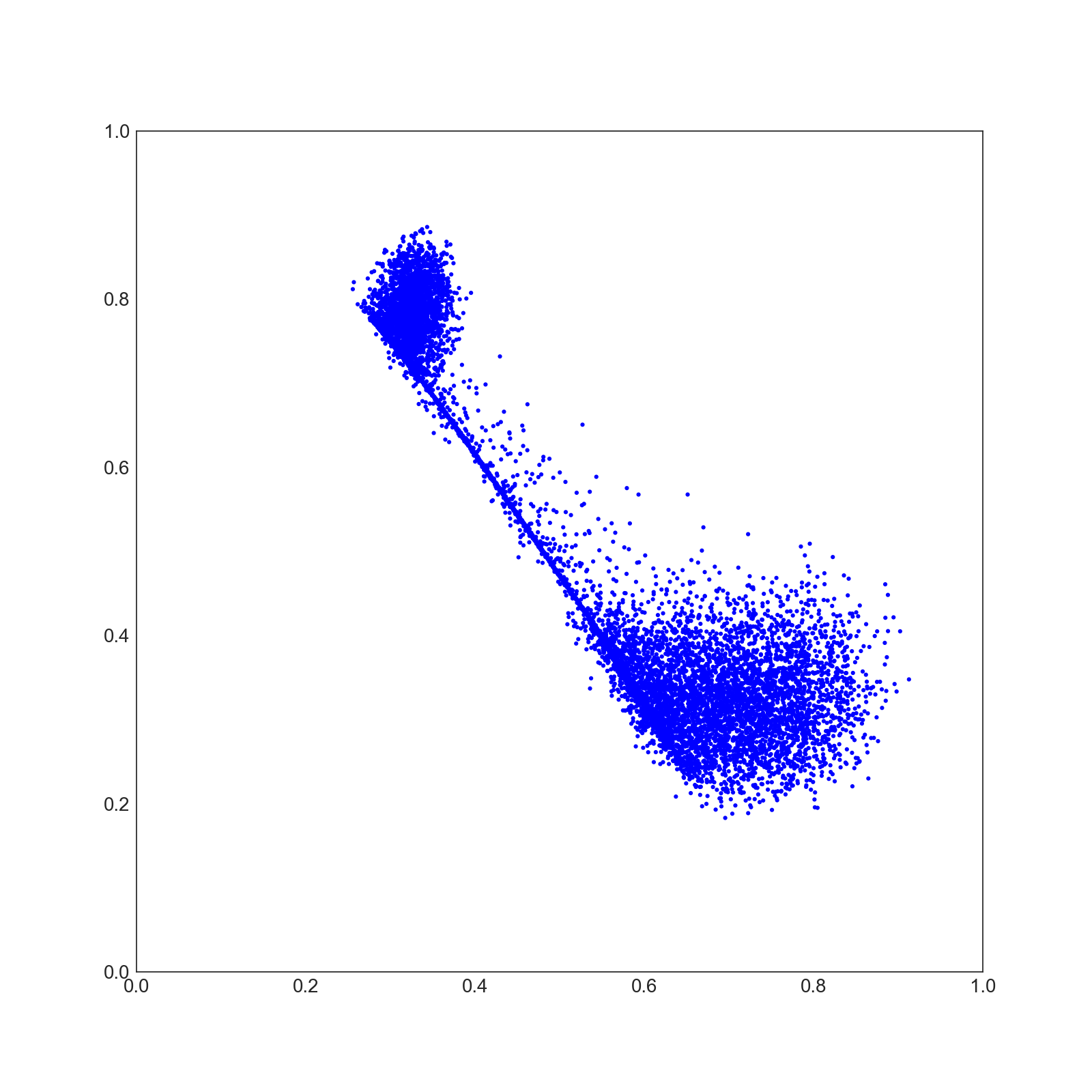}
  &\includegraphics[width=0.95cm, height=0.95cm]{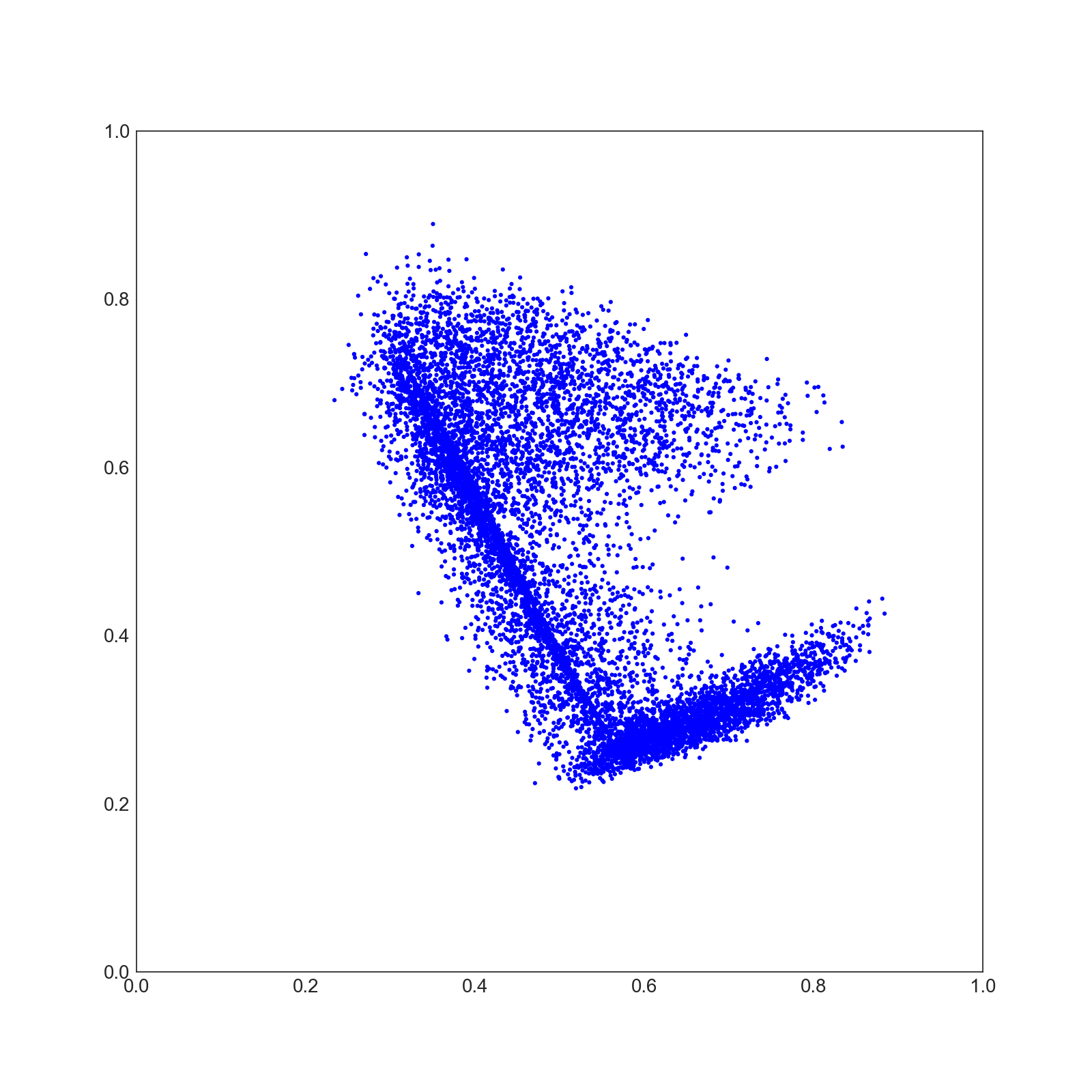}
  &\includegraphics[width=0.95cm, height=0.95cm]{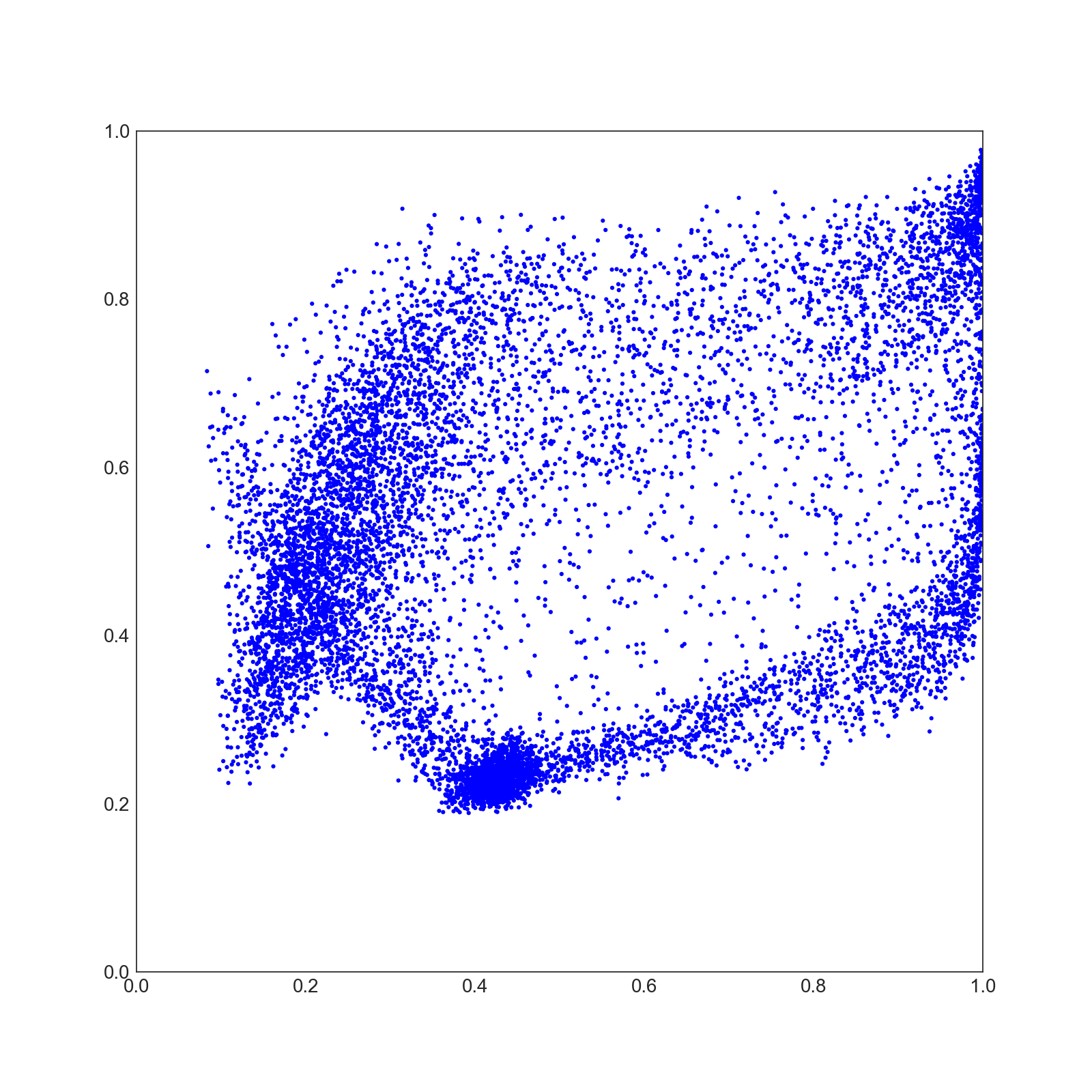}
  &\includegraphics[width=0.95cm, height=0.95cm]{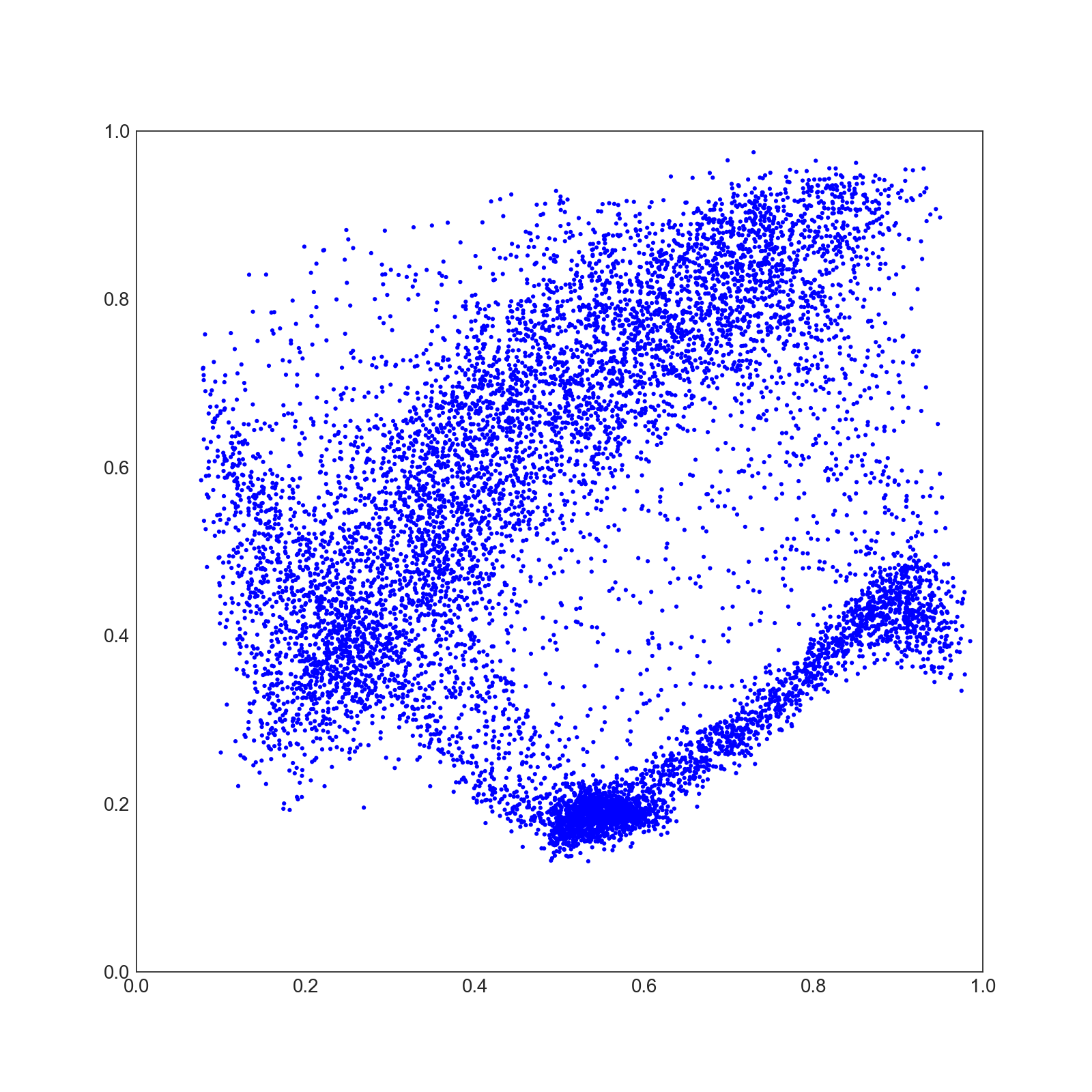}
  &\includegraphics[width=0.95cm, height=0.95cm]{scatter_true}\\
  
  
  {\footnotesize GAN$^{(2)}$}
  &\includegraphics[width=0.95cm, height=0.95cm]{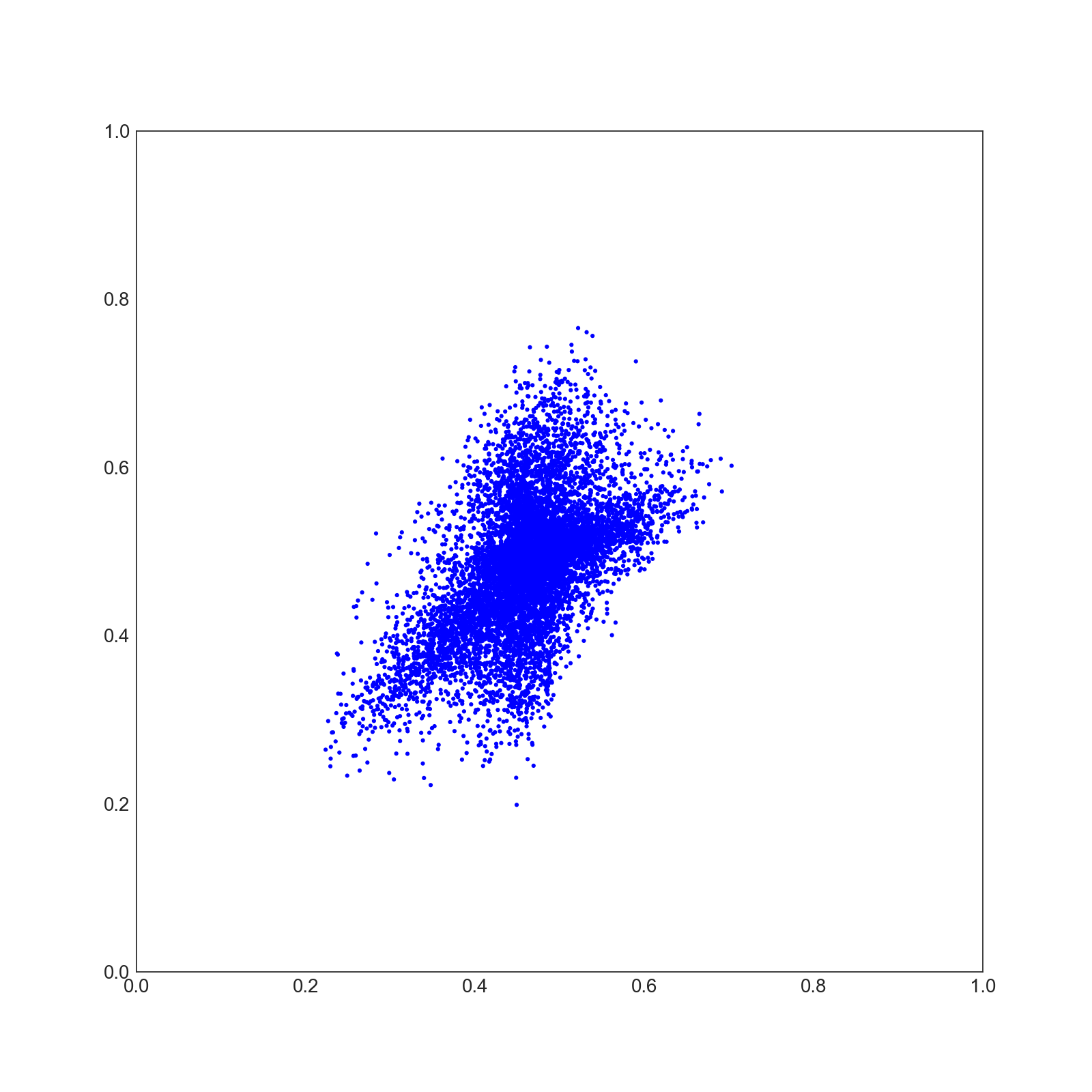}
  &\includegraphics[width=0.95cm, height=0.95cm]{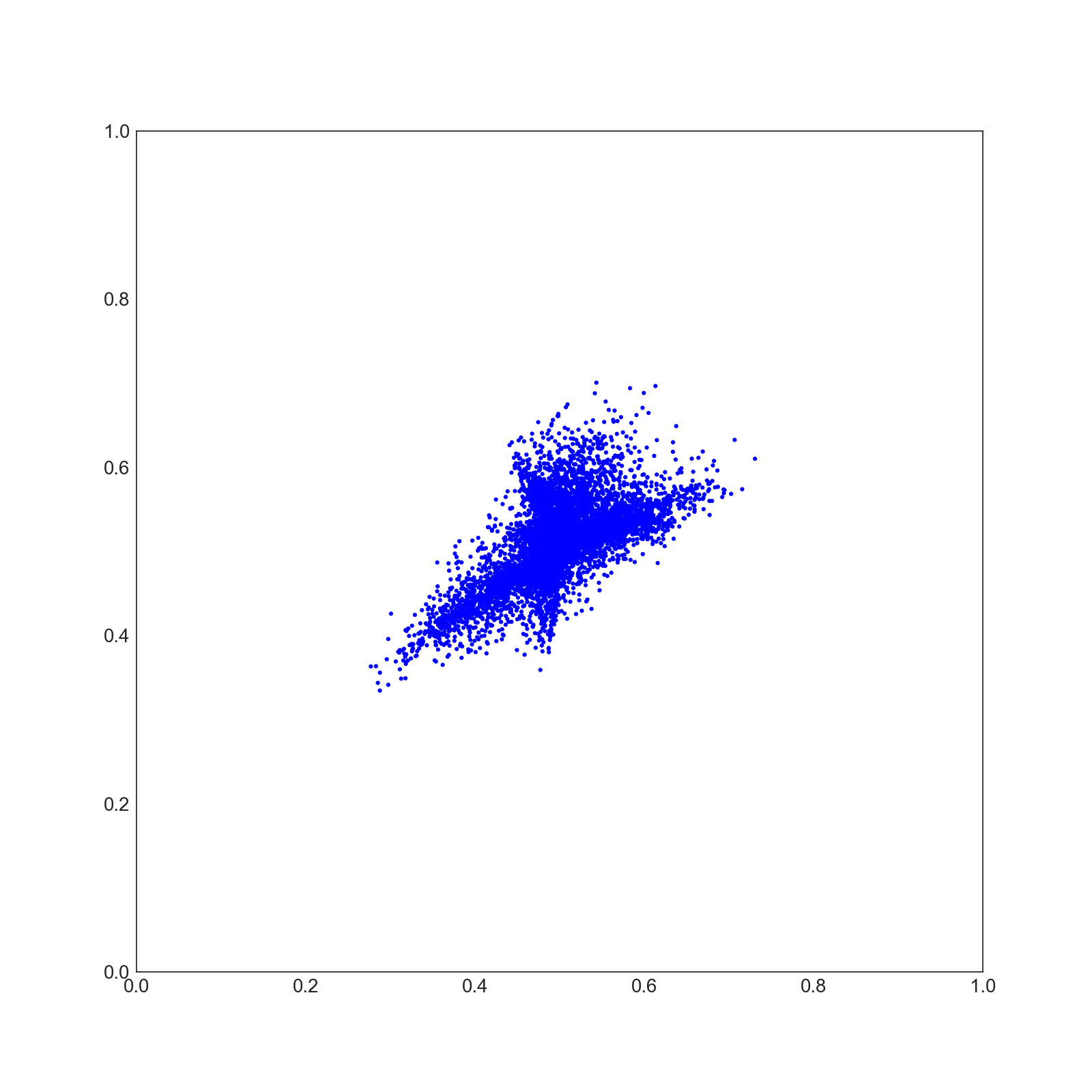}
  &\includegraphics[width=0.95cm, height=0.95cm]{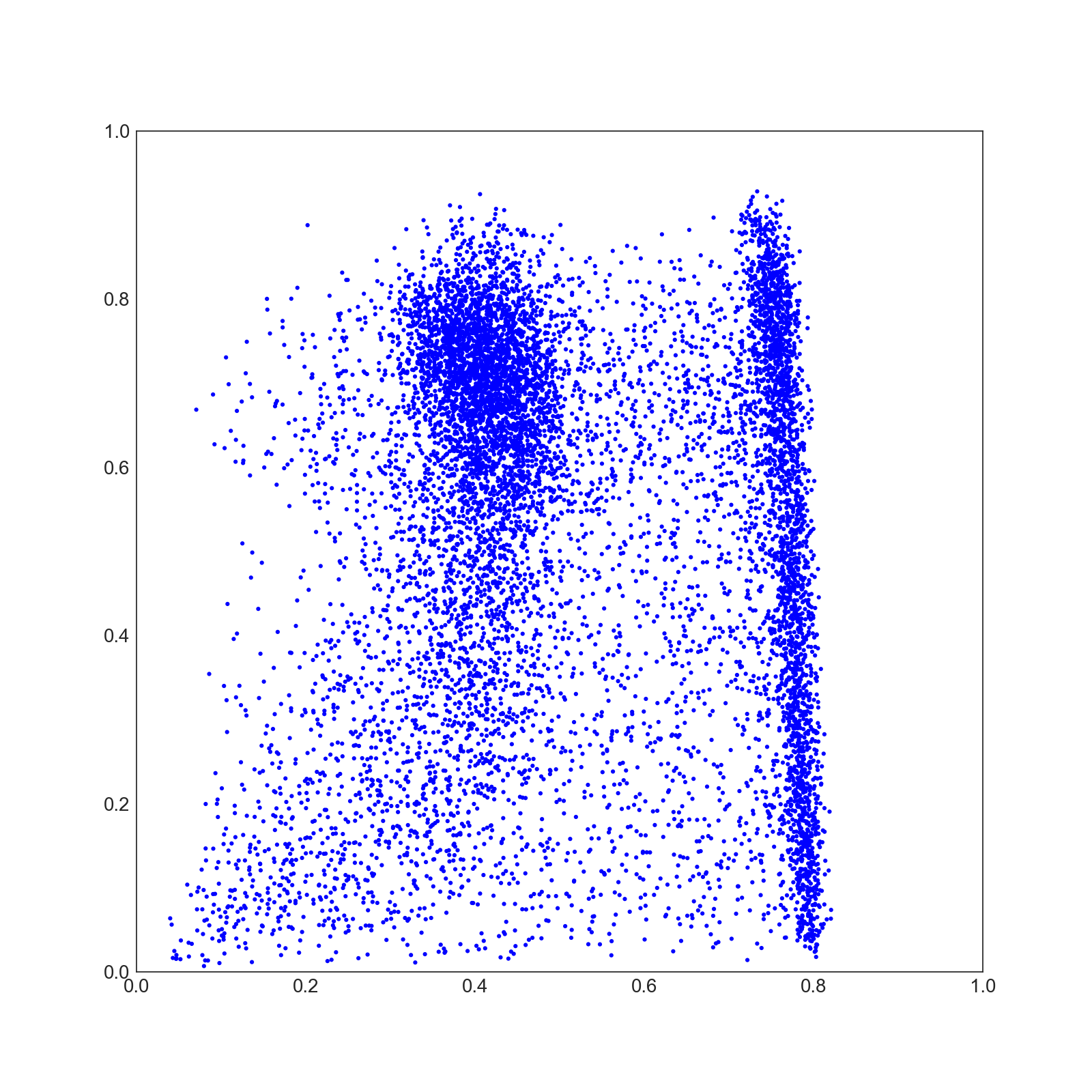}
  &\includegraphics[width=0.95cm, height=0.95cm]{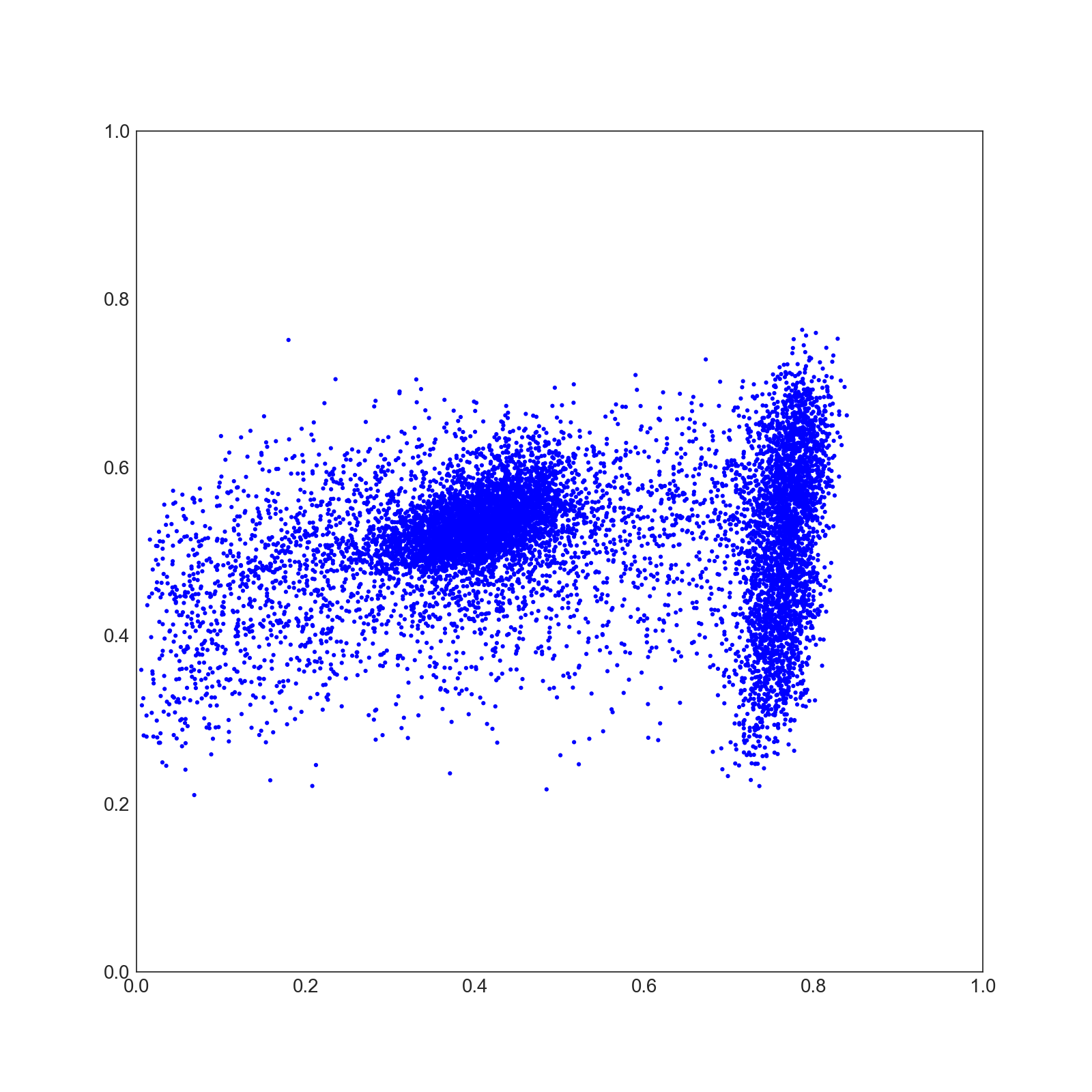}
  &\includegraphics[width=0.95cm, height=0.95cm]{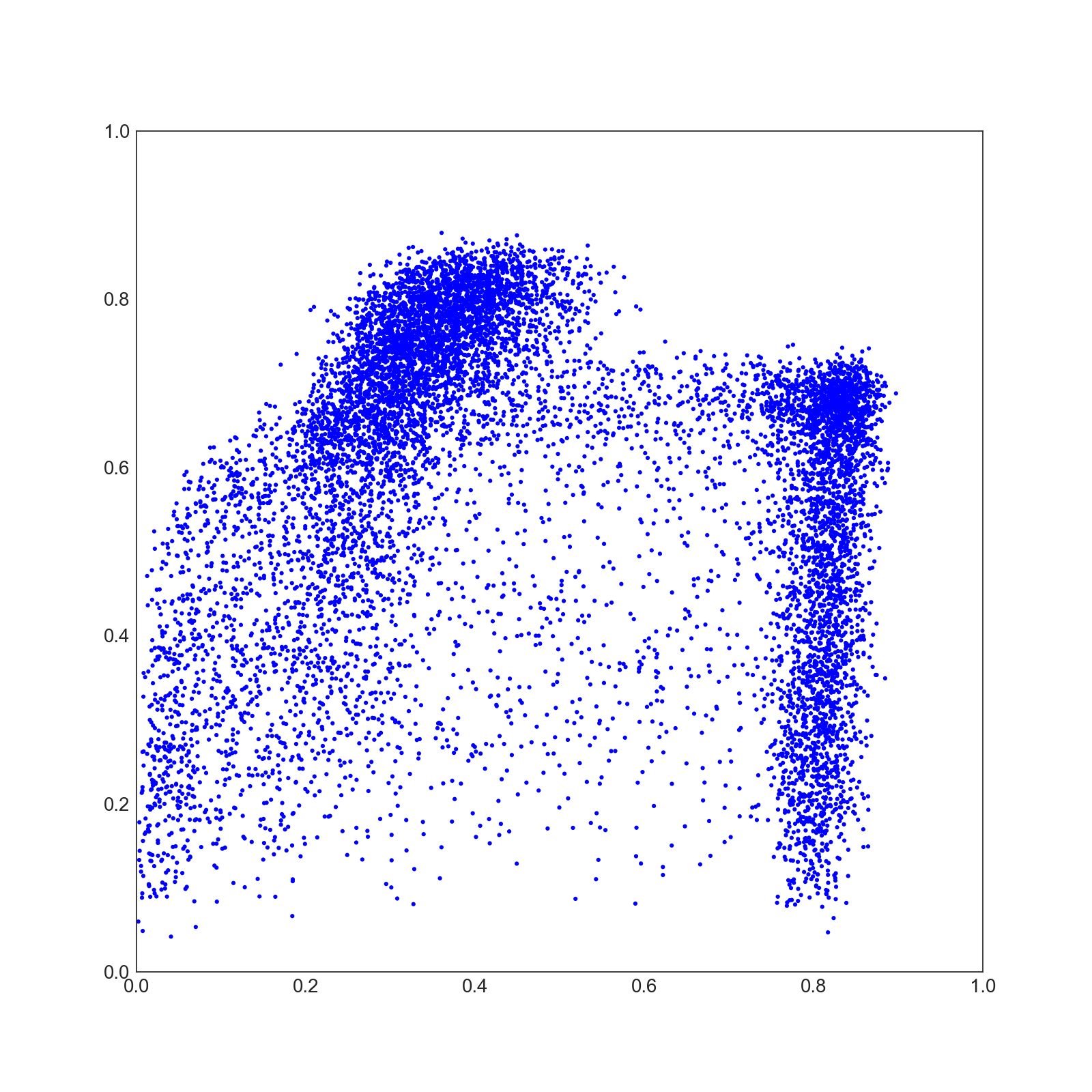}
  &\includegraphics[width=0.95cm, height=0.95cm]{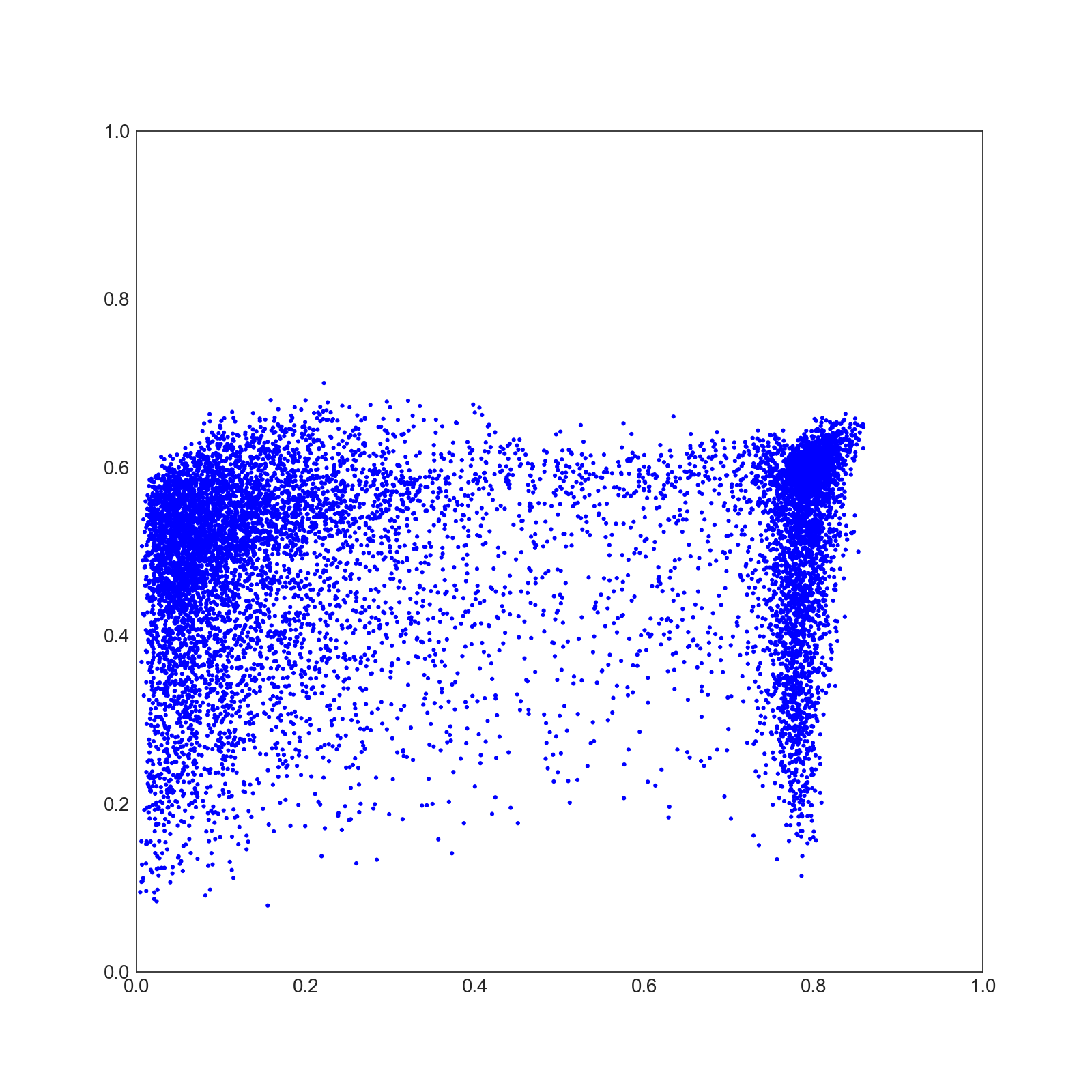}
  &\includegraphics[width=0.95cm, height=0.95cm]{scatter_true}\\


  {\footnotesize GAN$^{(3)}$}
  &\includegraphics[width=0.95cm, height=0.95cm]{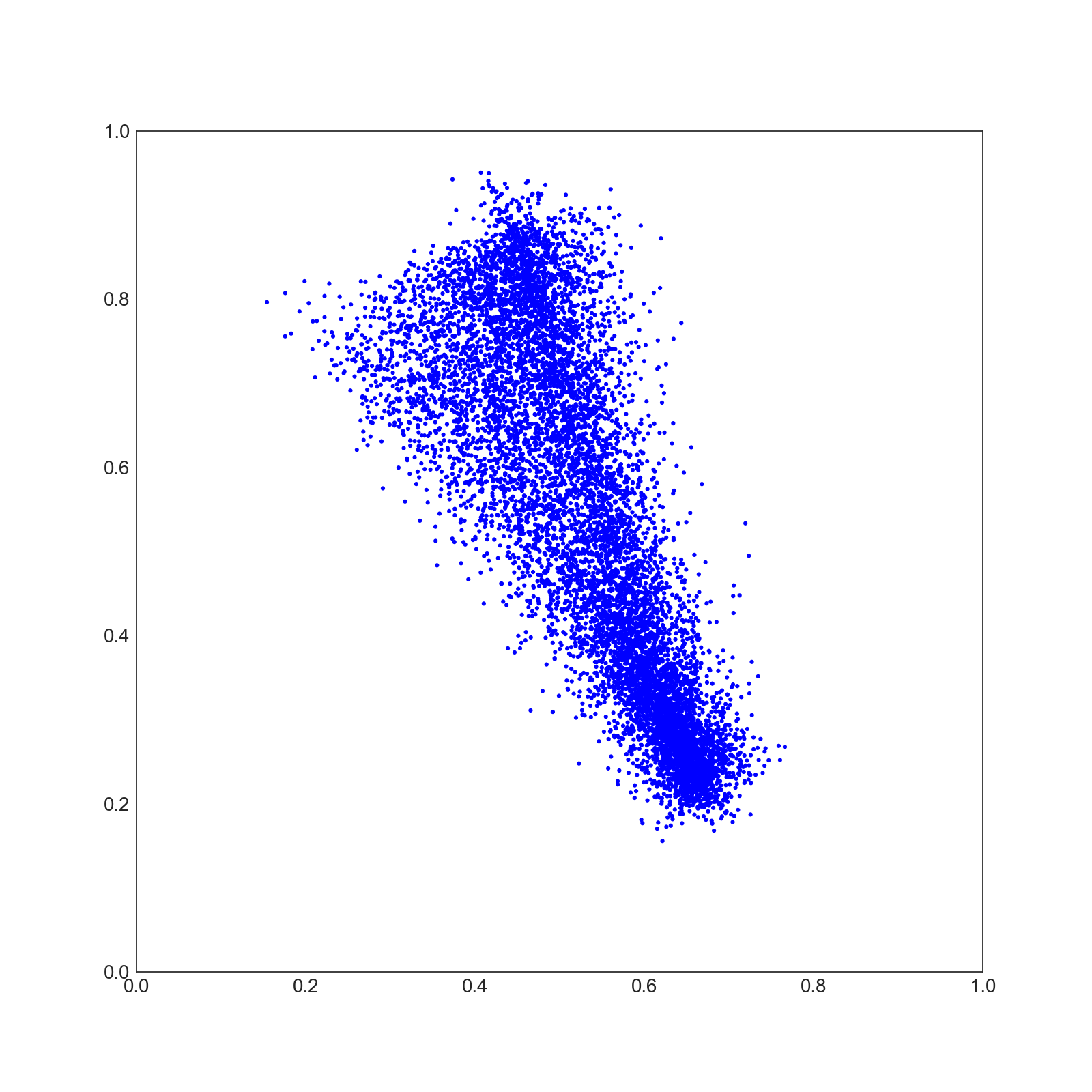}
  &\includegraphics[width=0.95cm, height=0.95cm]{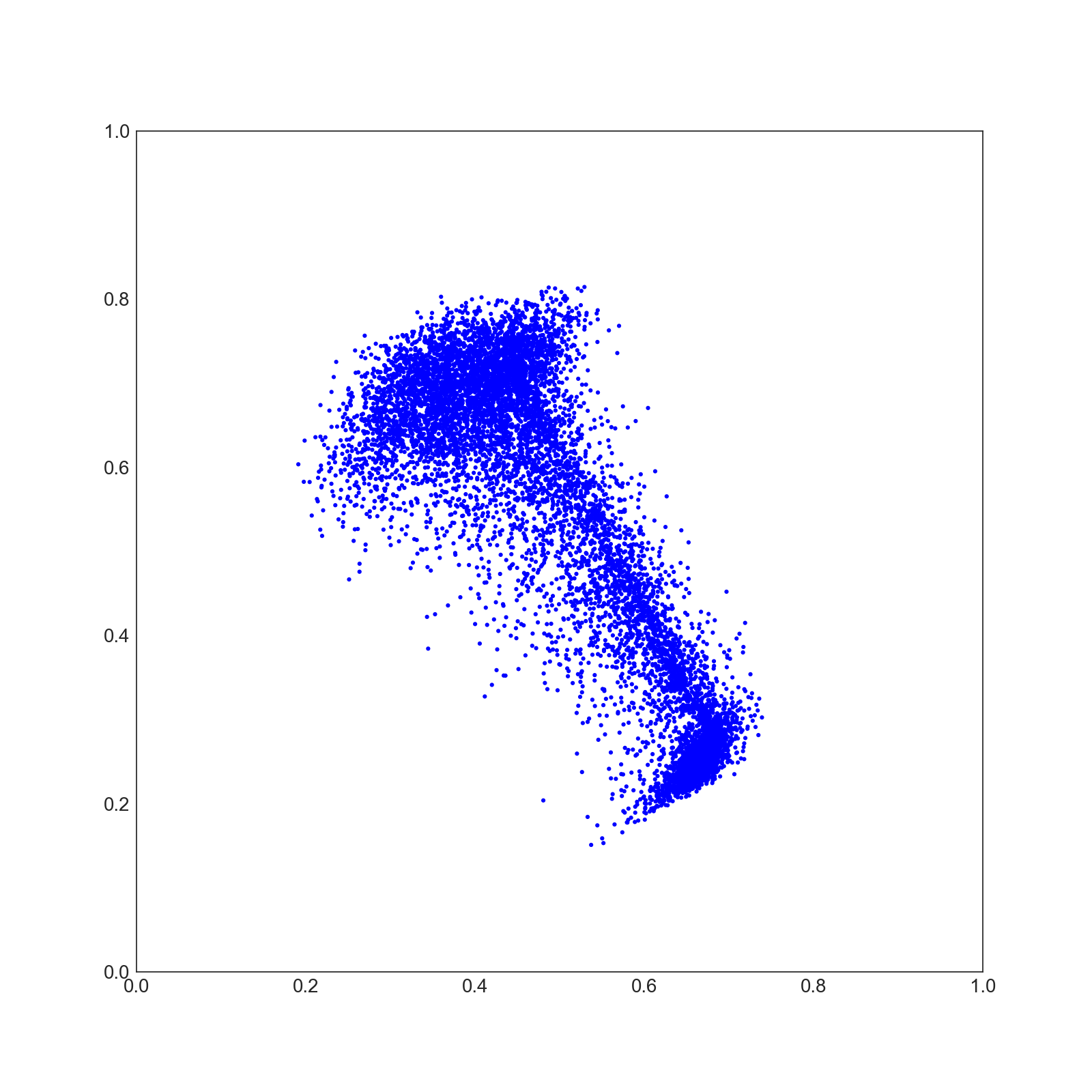}
  &\includegraphics[width=0.95cm, height=0.95cm]{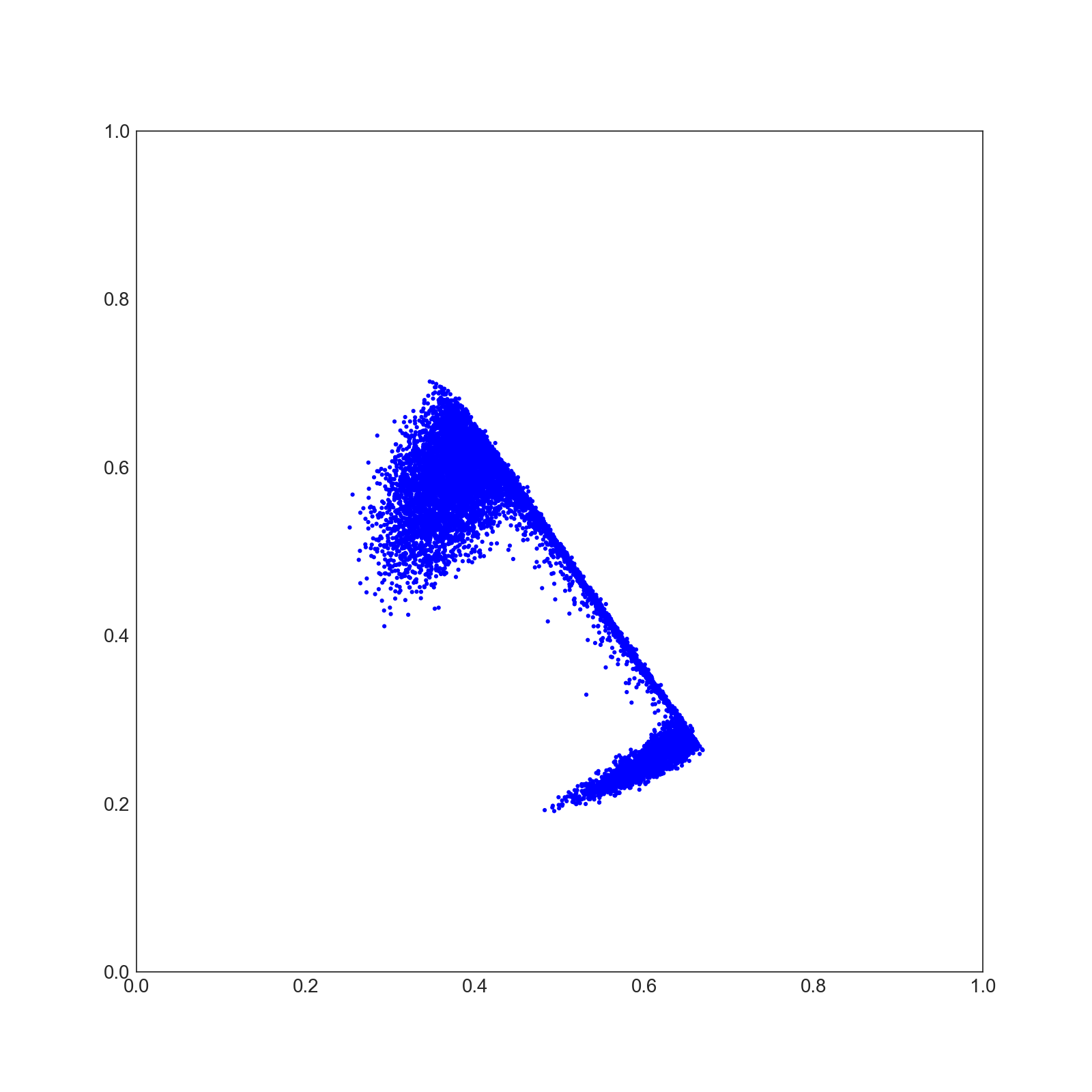}
  &\includegraphics[width=0.95cm, height=0.95cm]{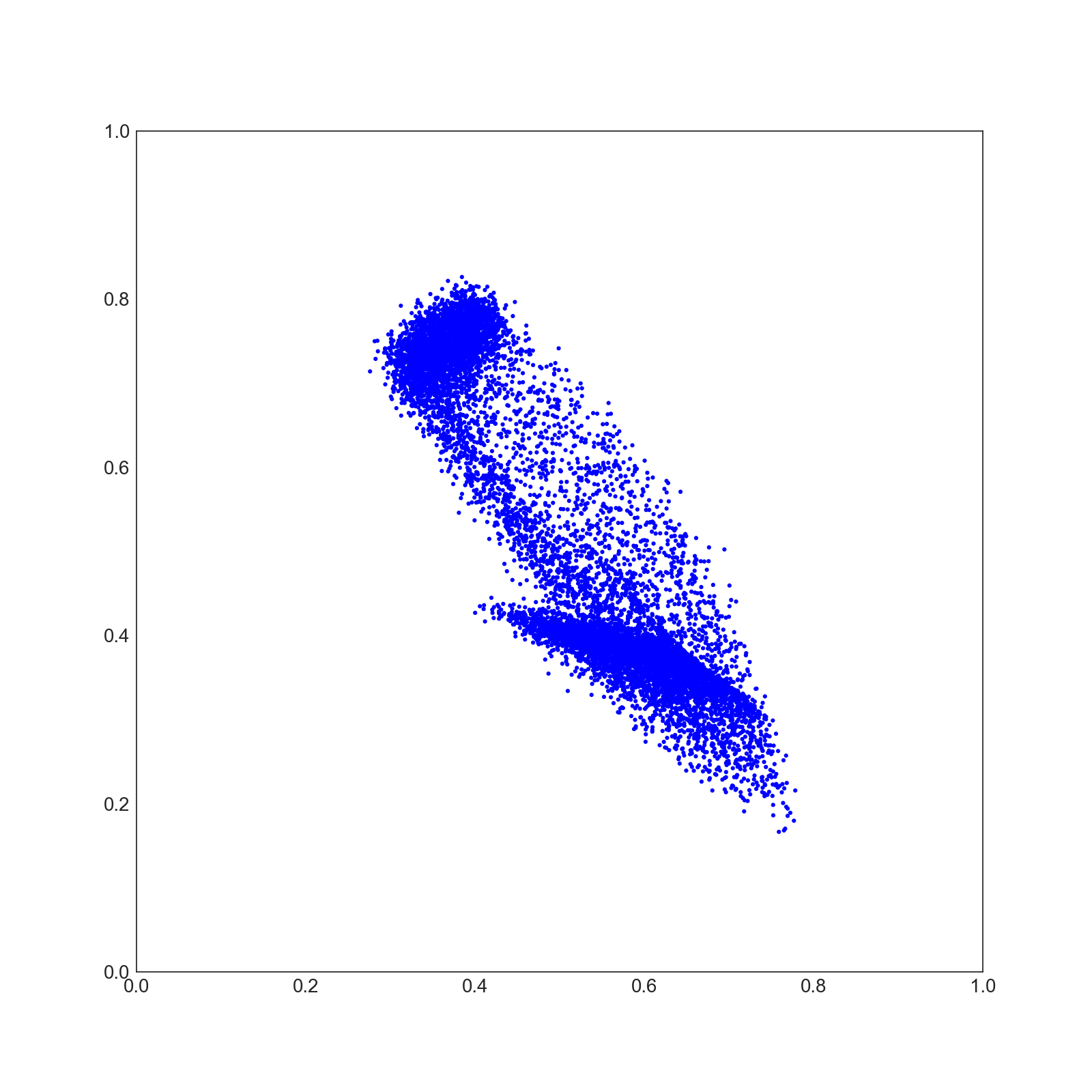}
  &\includegraphics[width=0.95cm, height=0.95cm]{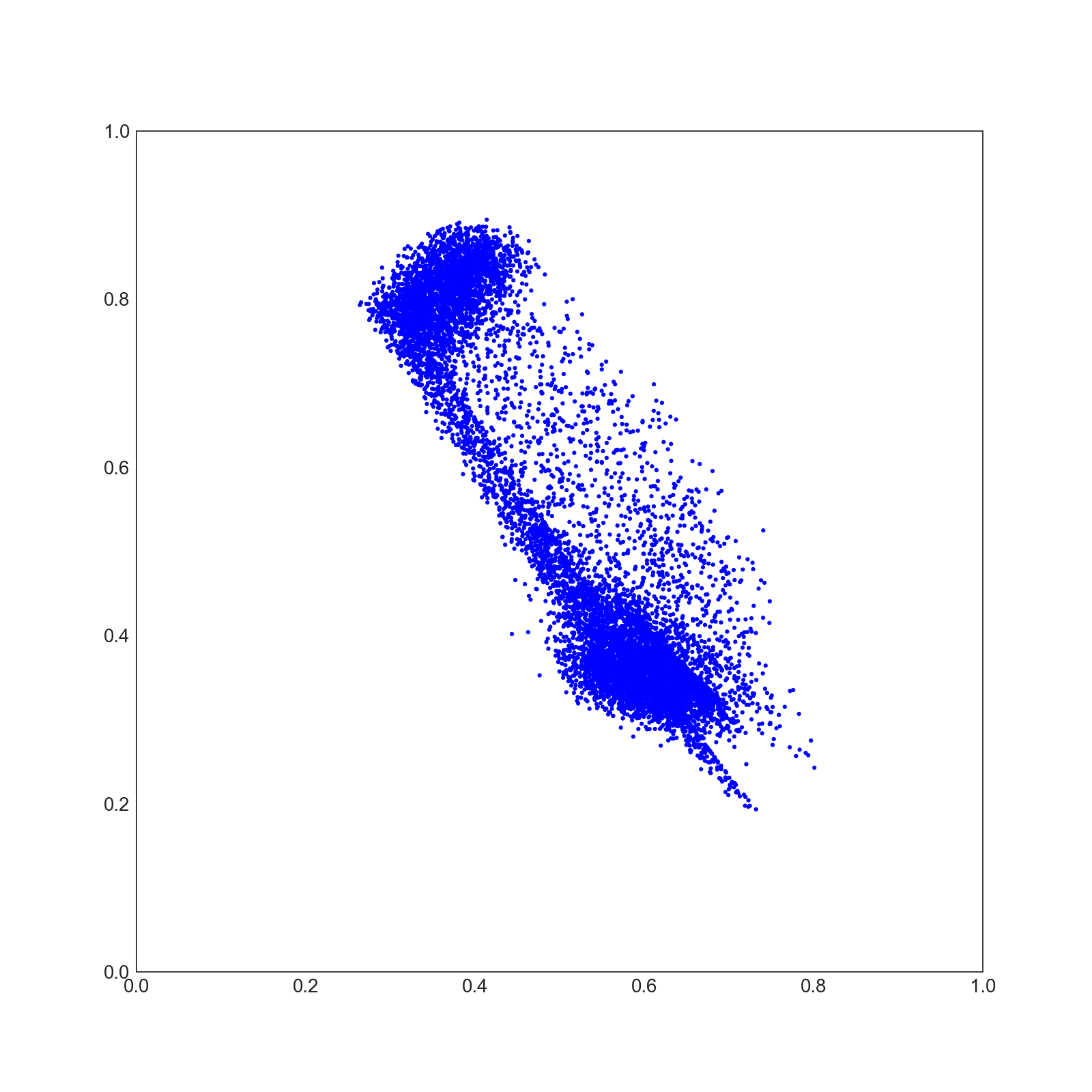}
  &\includegraphics[width=0.95cm, height=0.95cm]{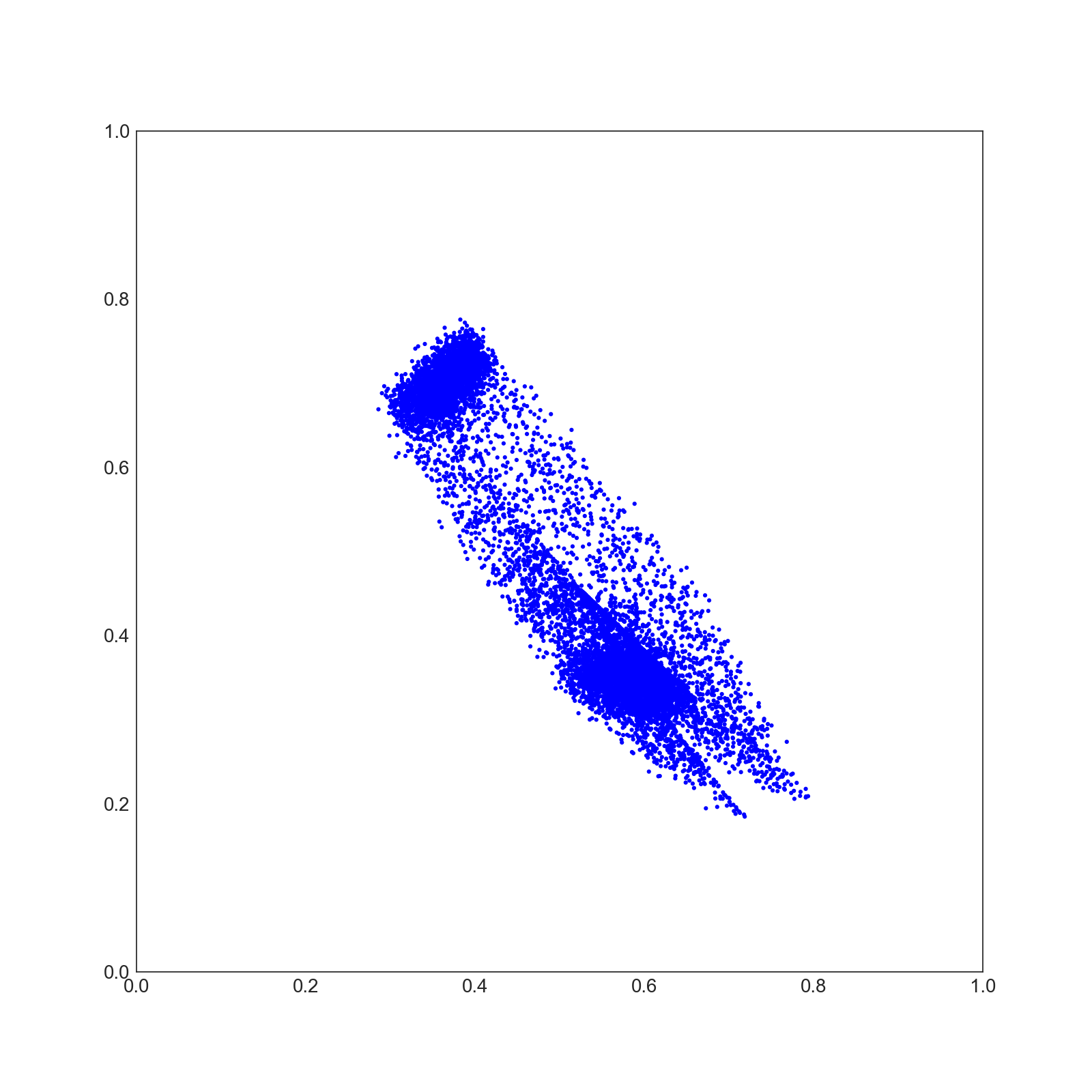}
  &\includegraphics[width=0.95cm, height=0.95cm]{scatter_true}\\
  
 \hline
  
  \end{tabular}
  \caption{ \footnotesize Samples from TGAN and three GANs with different size of layers. Each blue-colored figure represents scatter plot of $10,000$ samples generated at each training stage. It is observable that our model (TGAN) learns the distribution quickly while GANs struggle to do so. We set the prior to $z\sim \mathcal{U}(-1,1)$ for every synthetic experiment.}\label{fig:synthetic}
\end{figure}

In this section, we empirically compare the performance of TGAN and GANs using synthetic data; bivariate normal distributions. In particular, we collected $10,000$ data points sampled from $6$ clusters located circularly around the point $(0.5, 0.5)$. Each cluster is populated by bivariate normal distribution with $\sigma ^2 = 0.2$. We note that all models (\emph{e.g.} TGAN and GANs) employ approximately same number of trainable units. Figure~\ref{fig:synthetic} represents samples populated by TGAN and three GANs at each different training step. The experimental outcome demonstrates that; 1) our model converges to original distribution more quickly than GANs and 2) our model successfully learns the true distribution while others struggle to do so.

\section{Conclusion}\label{sc:conclusion}
The trend of employing models with a large number of parameters for generative adversarial learnings prohibits its applicability to systems with limited computational resources. At the same time, traditional affine transformations employed to GANs could significantly increase the model complexity. We present a new generative adversarial model with tensorized structure, TGAN. Two major advantages of TGAN are; 1) significant reduction in consumption of computational resource and 2) capability to capture multi-modal structure of datasets. We empirically demonstrated the compressive rate of $40$ times when compared to GAN while having negligible impact on the quality of generated samples. One of the future works we consider is to apply various tensor decomposition algorithms such as \cite{oseledets2011tensor, zhao2016tensor} to each tensor layer for further reduction in the number of model parameters.


\captionsetup[subfloat]{labelformat=empty}
\def\arraystretch{1}\tabcolsep=0.1pt

\bibliographystyle{IEEEbib}
\bibliography{Template}

\end{document}